\documentclass{article}

\usepackage{authblk}

\usepackage{graphicx,pdfpages}
\usepackage{booktabs}
\usepackage{multirow}
\usepackage[letterpaper,top=2cm,bottom=2cm,left=2.5cm,right=2.5cm,marginparwidth=1.75cm]{geometry}

\usepackage{amsmath,amssymb, bm}
\usepackage{graphicx}

\usepackage{algpseudocode}
\usepackage{algorithm}
\usepackage[colorlinks=true, allcolors=blue]{hyperref}

\usepackage{xcolor} 
\usepackage{comment}

\begin{document}


\title{Scalable Training of Continuous-Time Spiking Neural Networks with Differentiable Spike-Time Discretization}

\author[1,2]{Yusuke Sakemi}
\author[1]{Tomoya Takeuchi}
\author[3]{Takeo Hosomi} 
\author[1,2]{Kazuyuki Aihara}
\affil[1]{Research Center for Mathematical Engineering, Chiba Institute of Technology, Narashino, Japan}
\affil[2]{International Research Center for Neurointelligence (WPI-IRCN), The University of Tokyo, Tokyo, Japan}
\affil[3]{NEC Corporation, Kawasaki, Japan}

\date{\today}

\maketitle

\begin{abstract}
Continuous-time spiking neural networks (SNNs) provide an event-driven framework for temporal computation, computational neuroscience, and neuromorphic hardware. However, training deep continuous-time SNNs is severely constrained by the memory required for exact spike-time computation, which evaluates and retains candidate firing times over intervals determined by presynaptic spike ordering. Here we introduce a memory-efficient training framework based on differentiable spike-time discretization (DSTD) for leaky integrate-and-fire neurons with general membrane and synaptic time constants. DSTD maps irregular presynaptic spikes onto differentiable weighted events at fixed time points, replacing the input-dependent candidate dimension with $M$ fixed time intervals while accurately approximating continuous-time membrane-potential dynamics. This reduces candidate-related activation memory from $O(N_{\mathrm{out}}N_{\mathrm{in}})$ to $O(N_{\mathrm{out}}M)$ in the case of time-to-first-spike (TTFS) coding, where $N_{\mathrm{in}}$ and $N_{\mathrm{out}}$ denote the numbers of presynaptic and postsynaptic neurons, respectively.  
We further introduce synfire-chain-inspired temporal regularization that organizes layer-wise firing windows, mitigates dead-neuron failures, and enables pipeline-like processing. In dense LIF layers, DSTD reduced peak memory consumption by up to approximately 100-fold and training time by up to approximately 20-fold compared with exact spike-time computation. Together, these methods allowed us to train 9-layer convolutional SNNs on CIFAR-10 and 20-layer convolutional SNNs on Fashion-MNIST on a single GPU.
\end{abstract}



\section{Introduction}

Spiking neural networks provide a computational framework in which information is represented by discrete spike events and their timing \cite{Roy2019towards}. 
This makes them relevant both for understanding temporal coding in biological neural systems \cite{Yuste2024Neuronal} and for implementing event-driven, energy-efficient computation \cite{Kudithipudi2025neuromorphic}. 
Continuous-time SNNs are particularly attractive because they treat spike times as explicit continuous variables, rather than as events constrained to discrete simulation steps. However, despite this conceptual appeal, training continuous-time SNNs at the scale of modern deep architectures remains challenging.

Most scalable training methods for SNNs rely on surrogate gradients \cite{Neftci2019surrogate}. 
These methods have enabled deep SNNs by discretizing time and replacing the derivative of spike generation with a smooth or straight-through approximation. However, because they optimize time-discretized dynamics, they are not designed to directly control exact spike times in continuous time. 
This limitation is important when the objective is to study temporal coding or to map the learned dynamics to continuous-time analog neuromorphic hardware.

A more direct approach is to train continuous-time SNNs using analytical spike-time gradients. 
In particular, time-to-first-spike (TTFS) coding provides a simple framework in which each neuron emits at most one spike and information is represented by the first spike time  \cite{Bohte2002error,Montigny2016analytical,Mostafa2018supervised,Comsa2021temporal,Goltz2021fast,Sakemi2023supervised,Zhang2021rectified}. 
Such models allow exact or analytically tractable gradients and have been extended to several neuron and synapse models. However, their application to deep and wide networks remains limited.

Three obstacles are central to this limitation. 
First, in LIF neurons, the postsynaptic spike time depends on the ordering between presynaptic spikes and the threshold-crossing time. 
Therefore, the number of candidate spike times grows with the number of input spikes. 
Second, in deep TTFS-SNNs, neurons in the hidden layers may fail to fire, which blocks temporal information propagation and stalls the gradient flow (the so-called dead-neuron problem) \cite{Mostafa2018supervised,Comsa2021temporal,Sakemi2023supervised}.  
Third, because spike times propagate layer by layer, deeper SNNs suffer from reduced throughput, especially when implemented in neuromorphic  hardware.  

Here we address these obstacles by extending differentiable spike-time discretization (DSTD) \cite{Sakemi2025harnessing} to LIF neurons and combining it with temporal regularization inspired by synfire-chain dynamics \cite{Abeles1991corticonics, Diesmann1999stable,Ikegaya2004synfire,Zheng2014robust,Moldakarimov2015feedback}. 
DSTD maps irregular continuous-time spike trains to weighted spike events on fixed time points, reducing the number of candidate spike times from the number of presynaptic spikes to the number of DSTD steps. 
The temporal regularization encourages neurons in each layer to fire within prescribed temporal windows, producing temporally shifted spike distributions across layers and enabling pipeline-like processing. 
We refer to the resulting model as a synfire-chain SNN (Syn-SNN).

We validate the proposed framework in both single-neuron and deep-network settings. 
For LIF neurons receiving spike inputs, DSTD accurately approximates membrane-potential trajectories and spike times while substantially reducing computation time and memory consumption. 
In our benchmarks, these improvements included nearly two orders of magnitude lower peak memory consumption for dense presynaptic spike inputs. 
We further demonstrate that DSTD enables the training of a 9-layer convolutional continuous-time SNN on CIFAR-10, where the learned models exhibit layer-wise spike-time propagation under constrained temporal windows. 
Finally, to demonstrate scalability to substantially deeper architectures, we train a 20-layer convolutional Syn-SNN on Fashion-MNIST and show that organized spike propagation and pipeline-like processing are maintained throughout the network.

\section{Method}

\begin{figure*}
\centering
\includegraphics[clip, width=1.0\textwidth]{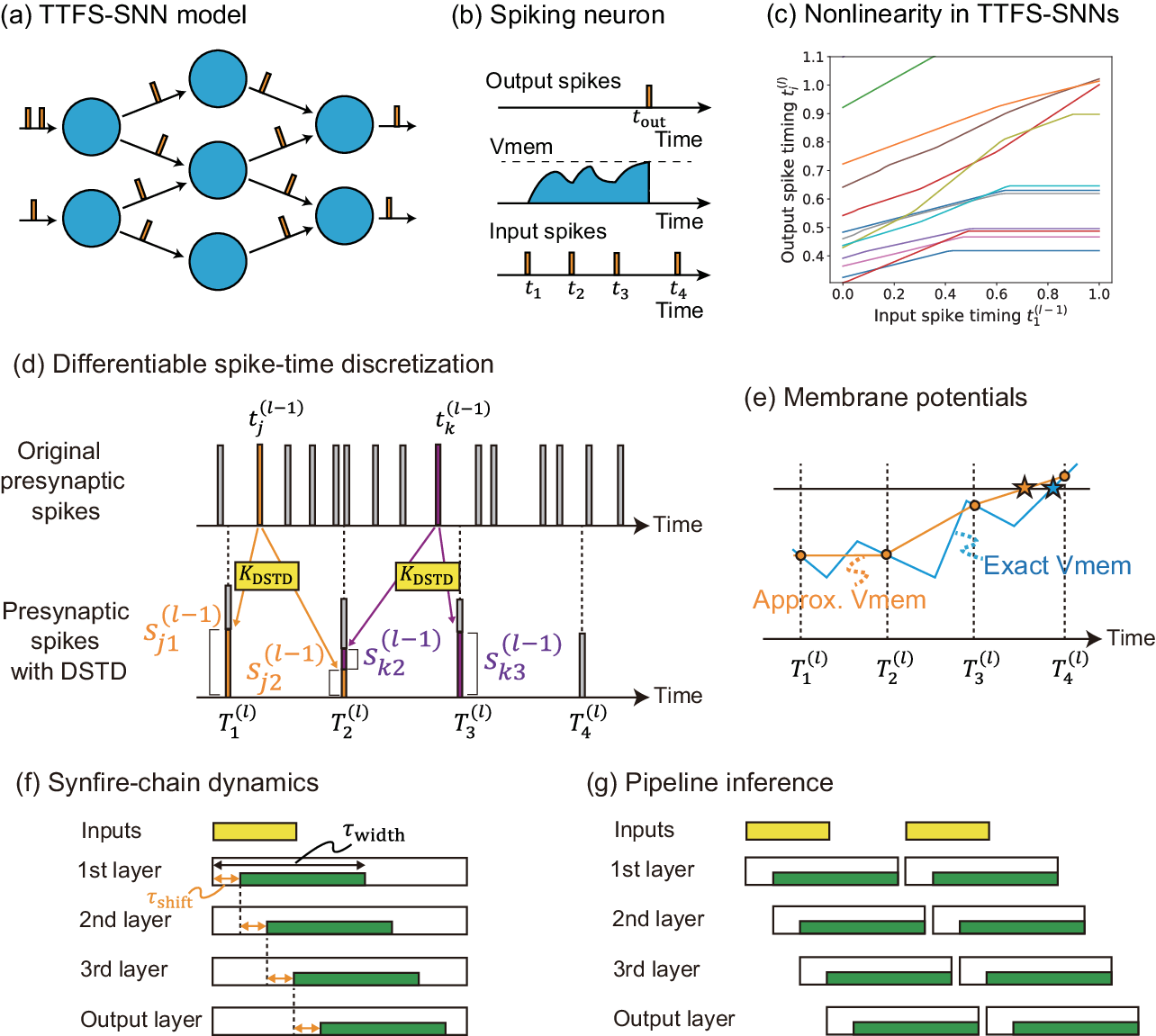}
\caption{{\textbf{Overview of DSTD for continuous-time TTFS-SNNs.}
(a) For simplicity, this study considers TTFS coding, in which each neuron is restricted to fire at most once. Extension to multiple spikes is formally possible in the same framework.
(b) Example dynamics of a LIF neuron.
(c) Change in the firing time $t_i^{(l)}$ of a non-leaky LIF neuron ($\tau_v=\tau_I=\infty$) when ten input spikes are provided and one of the input spike times, $t_1^{(l-1)}$, is varied. Here, $w_{i1}^{(l)}=1$.
(d) DSTD maps a spike train represented in continuous time to a spike train on discrete time points. For simplicity, the case in which all connection weights are one ($w_{ij}^{(l)}=1$) is shown.
(e) Owing to the kernel, the approximated membrane potential (the orange line) interpolates the exact membrane potential (the blue line).
(f) Learning synfire-chain dynamics. Neurons in each layer are allowed to operate only within a time window of width $\tau_\text{width}$, and the window is shifted by $\tau_\text{shift}$ across layers. The firing times in each layer are trained to fall within a time region of width $\tau_\text{width} - \tau_\text{shift}$. This is achieved by imposing a temporal penalty on each layer. 
(g) After the operation time for a given input datum has ended in each layer, the membrane potentials and synaptic currents are reset. The layer then starts operating in response to the next input datum, or another spike train. This mechanism enables pipeline operation in multilayer SNNs.} 
}
\label{fig:SNN}
\end{figure*}

In this study, we consider multilayer models composed of leaky integrate-and-fire (LIF) neurons (Figs. \ref{fig:SNN} (a) and (b)).
The LIF model is simple, yet it can reproduce the spiking behavior of biological neurons \cite{Gerstner2014neuronal}: 
\begin{align}
\begin{split}
\frac{dv_i^{(l)}}{dt} (t) &= -\frac{1}{\tau_v}v_i^{(l)}(t) + I_i^{(l)}(t), \\
    \frac{dI_i^{(l)}}{dt} (t) &= -\frac{1}{\tau_I}I_i^{(l)}(t) + u_i^{(l)}(t), \\
    u_i^{(l)}(t)&=\sum_{j=1}^{N^{(l-1)}} \sum_{q=1}^F w_{ij}^{(l)} \delta\left(t-t_{j,q}^{(l-1)}\right). 
\end{split} \label{eq:LIF_model}
\end{align}
Here, $v_i^{(l)}$ and $I_i^{(l)}$ denote the membrane potential and synaptic current of the $i$th neuron in the $l$th layer, respectively.
The parameters $\tau_v$ and $\tau_I$ denote the time constants of the membrane potential and synaptic current, respectively.
The term $u_i^{(l)}$ represents the spike input to this neuron, and $t_{j,q}^{(l-1)}$ denotes the $q$th firing time of neuron $j$ in the $(l-1)$th layer.
In time-to-first-spike (TTFS) coding, each neuron is restricted to fire at most once, namely $F=1$.
In this study, we adopt TTFS coding for simplicity and computational efficiency.
Therefore, hereafter, we omit the index for the spike count and use the simpler notation $t_j^{(l-1)}$.
$N^{(l)}$ is the number of neurons in the $l$th layer, and $w_{ij}^{(l)}$ is the connection weight from neuron $j$ in the $(l-1)$th layer to neuron $i$ in the $l$th layer.
Note that the method proposed in this study is also formally applicable to the multiple-spike case (see \ref{ss:multi-spikes} for details).
The firing time is defined as the first time at which the membrane potential $v_i^{(l)}(t)$ reaches the firing threshold $V_\text{th}$, that is, $v_i^{(l)}(t_i^{(l)}) = V_\text{th}$.

TTFS-SNNs have a unique mode of information processing, in which nonlinearity arises from the order of spikes \cite{Sakemi2023supervised}.
This property is particularly clear when $\tau_v=\tau_I=\infty$.
In this case, the firing time is given by \cite{Sakemi2023supervised}
\begin{align}
t_i^{(l)} = \frac{V_\text{th} + \sum_{j\in \mathcal{S}_i} w_{ij}^{(l)} t_j^{(l-1)}}{\sum_{j\in \mathcal{S}_i} w_{ij}^{(l)}} .\label{eq:spike_no_leak}
\end{align}
Here, $\mathcal{S}_i$ is the index set of presynaptic spikes that arrive before neuron $i$ in the $l$th layer fires.
In TTFS-SNNs, if the input spikes to the network are denoted by $t^{(0)}$, information can be regarded as being transmitted through layers as
$t^{(0)}\longrightarrow t^{(1)} \longrightarrow \cdots \longrightarrow t^{(L)}$.
Importantly, Eq. (\ref{eq:spike_no_leak}) is linear with respect to the input spike times $t_j^{(l-1)}$ for a fixed index set $\mathcal{S}_i$, but nonlinearity is introduced when the index set $\mathcal{S}_i$ changes \cite{Sakemi2023supervised}.
An example of the change in the firing time when one input spike time $t_1^{(l-1)}$ is varied among ten input spikes is shown in Fig. \ref{fig:SNN} (c).

In this way, TTFS-SNNs realize nonlinearity through the temporal order of spikes and constitute a form of information processing that differs from artificial neural networks (ANNs) based on, for example, ReLU neurons \cite{Nair2010rectified}.
Therefore, unlike models that assume firing after all inputs have been received \cite{Rueckauer2018conversion} or models that separate the input and output phases \cite{Zhang2019tdsnn,Park2020t2fsnn,Stockl2021,Sakemi2022spiking,Stanojevic2024high}, direct mapping from ANNs to SNNs is nontrivial.
Thus, SNNs must be trained directly.

To calculate the firing time of each layer in a TTFS-SNN
(Eq.~\eqref{eq:spike_no_leak} for $\tau_v=\tau_I=\infty$; see Appendix~\ref{ss:model} for other cases),
the presynaptic spikes are first ordered according to their arrival
times.
A candidate firing time is then evaluated for each interval between
successive presynaptic spikes, and the first candidate satisfying the
corresponding interval condition is selected as the output spike time.

The cumulative sums required to evaluate these candidates can be
computed efficiently using parallel prefix-scan operations
\cite{Zhou2021temporal,Yamamoto2024can,Sakemi2023sparse,
Morrill2026bullet,Che2026parallel}.
Nevertheless, the exact formulation produces
$N^{(l-1)}$ candidate intervals for every neuron in layer $l$.
During reverse-mode automatic differentiation, intermediate variables
associated with these candidates must be retained, resulting in a
candidate-related activation-memory requirement of
\[
O\!\left(N^{(l)}N^{(l-1)}\right).
\]
Moreover, the candidate dimension depends on the number and ordering of
the presynaptic spikes, leading to input-dependent sorting and
irregular tensor operations.
To reduce this burden, we introduce differentiable spike-time
discretization (DSTD).

As shown in Fig. \ref{fig:SNN} (d), DSTD approximates the input spike train $u_i^{(l)}{(t)}$ by a spike train with real-valued weights on predetermined discrete time points \cite{Sakemi2025harnessing}:
\begin{align}
\hat{u}_i^{(l)}(t) = \sum_{m=0}^{M} \hat{w}_{im}^{(l)} \delta\left(t-T_m^{(l)}\right). 
\end{align}
Here, $M$ is the number of discrete time steps, and $\hat{w}_{im}^{(l)}\in \mathbb{R}$ denotes the spike weight at time $T_m^{(l)}$, which is given as follows:
\begin{align}
    \hat{w}_{im}^{(l)} &=\sum_{j=1}^{N^{(l-1)}}  w_{ij}^{(l)} s_{jm}^{(l-1)}\\
    s_{jm}^{(l-1)} &=\begin{cases}
        K_\text{DSTD} \left( t_j^{(l-1)} - T_m^{(l)} \right), &\text{ when } T_{m-1}^{(l)}\le  t_j^{(l-1)} <T_{m+1}^{(l)}, \\
        0 &\text{ otherwise.}
        \end{cases} \label{eq:DSTD}
\end{align}

Here, $K_\text{DSTD}$ is a kernel function that maps a single spike with weight one to a spike train on discrete time points.
By adopting a differentiable function for this kernel, standard training by error backpropagation can be applied.
Previous work showed that, when the time constants of the LIF model satisfy $\tau_v=\tau_I=\infty$, the subthreshold membrane potential approximated by DSTD, $\hat{v}_i^{(l)}(t)$, exactly agrees with the exact membrane potential at the discrete time points, namely $\hat{v}_i^{(l)}(T_m^{(l)}) = v_i^{(l)}(T_m^{(l)})$ \cite{Sakemi2025harnessing}.
We newly extend the function $K_\text{DSTD}$ in Eq. \eqref{eq:DSTD} so that, for arbitrary finite time constants of LIF neurons $(\tau_v < \infty, \tau_I<\infty)$, the approximated membrane potential agrees with the exact solution at the discrete time points (see Appendix \ref{ss:DSTD} for details).
That is, as shown in Fig. \ref{fig:SNN} (e), the approximated membrane potential acts as an interpolation, enabling highly accurate approximation of both the membrane potential and the firing time.
Furthermore, by increasing the number of DSTD steps $M$, the approximated membrane potential converges to the exact solution (see Appendix \ref{ss:additional_experiments} for empirical results).

Using DSTD, the number of candidate intervals is reduced from the
number of presynaptic spikes, $N^{(l-1)}$, to the fixed number of DSTD
intervals, $M$.
Consequently, the candidate-related activation-memory requirement is
reduced from
\[
O\!\left(N^{(l)}N^{(l-1)}\right)
\]
to
\[
O\!\left(N^{(l)}M\right).
\]
When $M \ll N^{(l-1)}$, this reduction is substantial.
DSTD also eliminates spike-time sorting from the candidate evaluation
and replaces input-dependent candidate intervals with regular,
fixed-size tensor operations.
A detailed discussion of the computational and memory complexities is
provided in Appendix \ref{ss:complexity}.

Next, we solve the remaining problems in training TTFS-SNNs, namely the instability of learning caused by the dead neuron problem and the decrease in processing efficiency when the network is made deeper, by introducing synfire-chain dynamics.
As shown in Fig. \ref{fig:SNN} (f), this synfire-chain dynamics can be realized simply by applying a temporal penalty term \cite{Sakemi2023supervised} to each layer, which causes spikes in each layer to be distributed within a specific time window (see Appendix \ref{ss:learning_algorithm} for details).
Because the temporal penalty term encourages all neurons to fire, the dead neuron problem is suppressed.
Furthermore, as shown in Fig. \ref{fig:SNN} (g), pipeline operation becomes possible, in which the next datum can be input before the neurons in the final layer fire.
Thus, a decrease in data-processing throughput due to multilayering can be prevented. 
We refer to a continuous-time TTFS-SNN equipped with this synfire-chain temporal regularization as a synfire-chain SNN (Syn-SNN). 
Throughout this paper, Syn-SNN denotes the network model, whereas DSTD denotes the training method used to compute its spike times efficiently; the two are independent and are combined in our experiments.

\begin{figure*}
\centering
\includegraphics[clip, width=\textwidth]{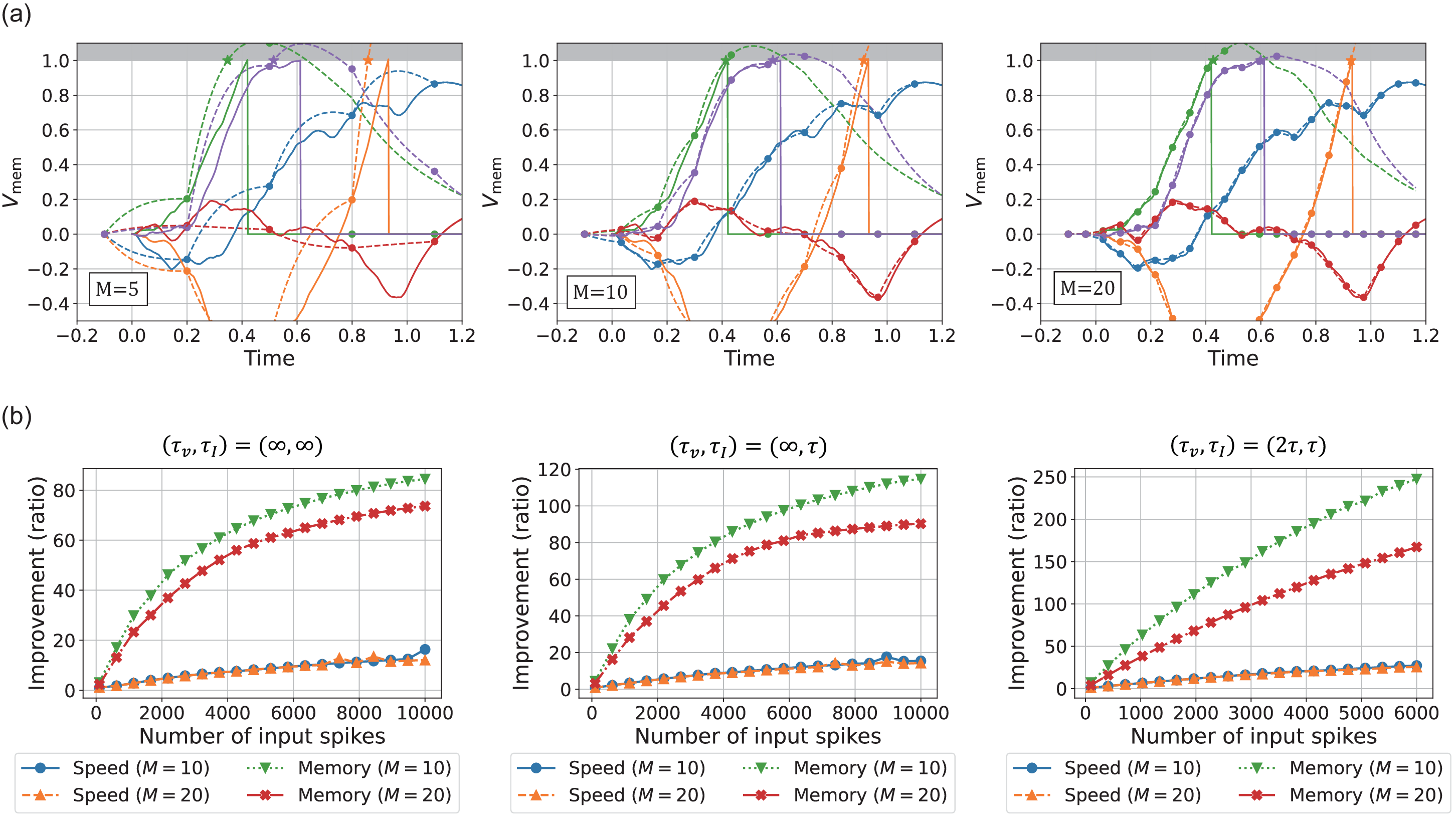}
\caption{\textbf{Simulation of LIF models using DSTD.} (a) Time evolution of 10 LIF neurons with time constants $\tau_v=0.4$  and $\tau_I=0.2$  when 1000 input spikes are randomly given in the interval $[0,1]$.
The solid line shows the exact membrane-potential waveform, the points and the dashed line show the membrane potentials calculated using DSTD.
(b) Improvement in computational efficiency when using DSTD. For LIF models with different time constants, input spikes were randomly given in the interval $[0,1]$ to a single-layer model consisting of 1000 neurons, and the processing time and peak memory were measured while varying the number of input spikes.
The batch size was set to 100, the number of data samples was 1000, and the time required for one epoch was measured.
}
\label{fig:DSTD_exp}
\end{figure*}

\section{Experiments}

DSTD achieves both high approximation accuracy and computational efficiency.
Figure \ref{fig:DSTD_exp} (a) shows the time evolution of the membrane potentials of 10 LIF neurons.
These membrane potentials were obtained by applying 1000 random input spikes in the time interval $[0,1]$ to LIF neurons with time constants $(\tau_v, \tau_I)=(0.4, 0.2)$.
The exact solutions are shown by solid lines, whereas the results obtained using DSTD are shown by dashed lines.
The spike times obtained using DSTD are indicated by star markers.
In the subthreshold regime, the membrane potentials approximated by DSTD at the discrete time points agree with the exact solutions.
In contrast, deviations in the membrane-potential calculation arise between adjacent discrete time points, resulting in small errors in the spike times.
However, these errors decrease as $M$ increases.
Indeed, when the number of DSTD steps is small, such as $M=5$, visible errors are observed, whereas for $M=20$, the errors are reduced to a negligible level.
Figure \ref{fig:DSTD_exp} (b) shows the results of the runtime and peak memory evaluation for a single-layer model consisting of 1000 LIF neurons.
In this experiment, random spikes with various input sizes were applied to the model in the interval $[0,1]$.
The batch size was set to 100, the number of data samples was 1000, and the model was trained for 5 epochs; we measured the average computation time per epoch and the peak memory consumption.
The computational-efficiency gains, measured as ratios relative to the exact spike-time computation method, are shown for $M=10$ and $M=20$.
We also show the results for three LIF models with $(\tau_v, \tau_I)=(\infty, \infty), (0.2, \infty), (0.4, 0.2)$.
For all LIF models and all values of $M$, the speedup and memory-efficiency gains increase as the number of input spikes increases.
This behavior arises because the exact formulation evaluates and retains
intermediate variables for a number of candidate intervals that grows
with the number of presynaptic spikes, whereas DSTD uses a fixed number
of intervals, $M$.
Consequently, when $N^{(l-1)} \gg M$, DSTD substantially reduces the
candidate-related tensor size and memory consumption.
In particular, the improvement in memory efficiency is substantial: even for $M=20$, which accurately approximates spike times, the memory efficiency improves by a factor of 60 to 150 when the number of inputs reaches 6000.

\begin{figure*}
\centering
\includegraphics[clip, width=0.9\textwidth]{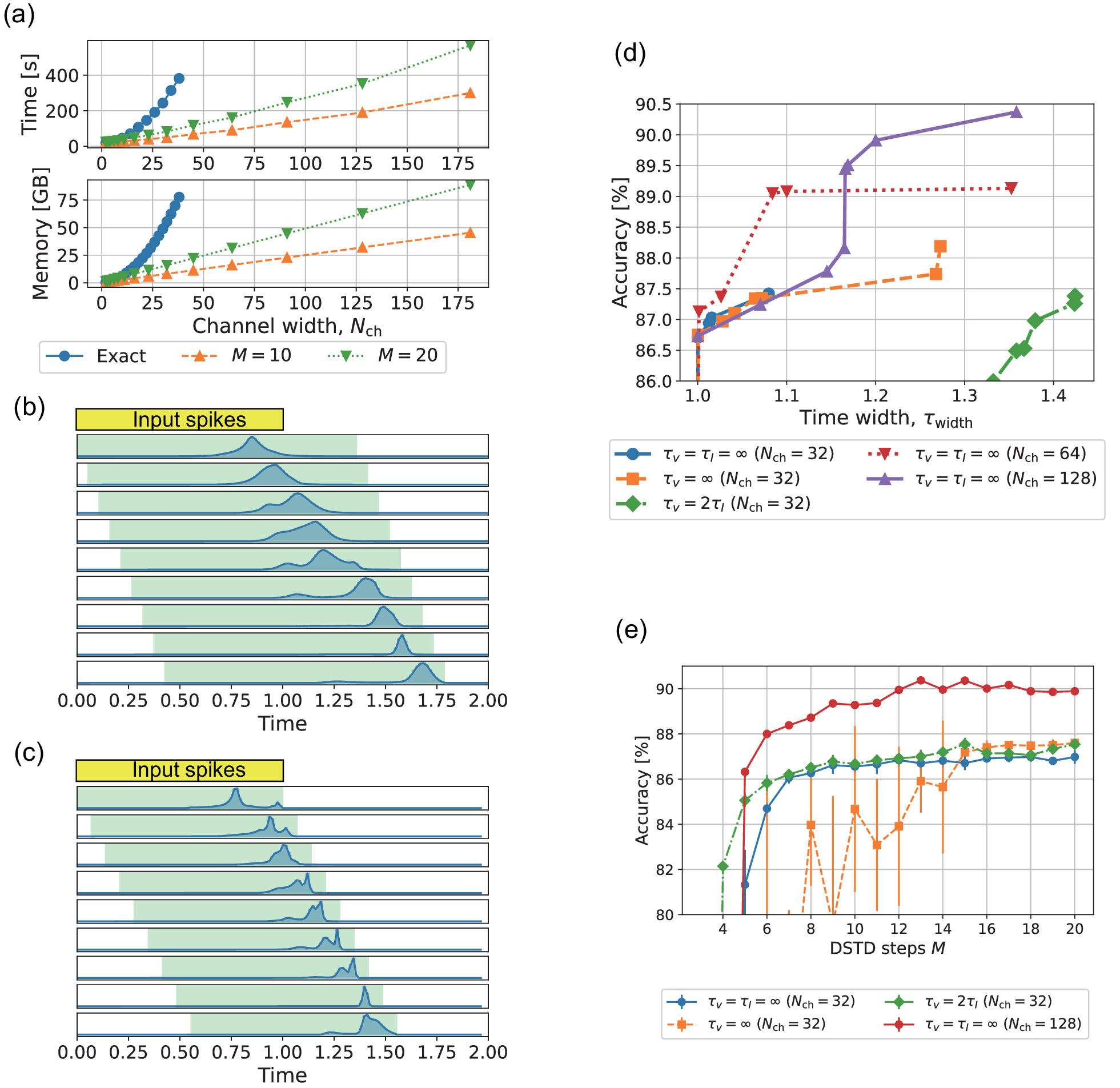}
\caption{\textbf{Training results for the 9-layer model (Syn-SNN-9).} 
(a) Runtime and memory consumption as functions of the width $N_\text{ch}$ of the network consisting of the LIF models with   $\tau_v=\tau_I=\infty$, comparing the cases with DSTD ($M=10$ and $M=20$) and without DSTD (Exact).
(b) Spike-time distributions of the trained Syn-SNN-9, shown from the first layer at the top to the second layer and subsequent layers below. The bottom panel corresponds to the output layer. The operating time-window width of each layer in this model is $\tau_\text{width}\approx 1.36$.
(c) Spike-time distributions of the 9-layer network for $\tau_\text{width}=1$.
(d) For three LIF models, Pareto fronts obtained by hyperparameter search based on Bayesian optimization are shown, where the optimization objectives are to maximize accuracy and minimize the time-window width $\tau_\text{width}$.
(e) Accuracy as a function of the number of DSTD steps $M$ used during training for different models. Error bars indicate the standard deviation over five trials. For the model with $N_\text{ch}=128$, only one trial was conducted for each value of $M$.
}
\label{fig:Pareto}
\end{figure*}

As a more practical network architecture, we trained a 9-layer convolutional Syn-SNN (Syn-SNN-9)  using DSTD.
We used the CIFAR-10 dataset \cite{CIFAR10}.
CIFAR-10 consists of 50000 natural images for training and 10000 natural images for testing.
The classification accuracy was evaluated on the test dataset, and in this evaluation we used a large number of DSTD steps ($M=40$) so that the continuous-time dynamics was accurately reproduced (see \ref{ss:DSTD_convergence} for the validity of this procedure). 
Details of the network architecture and training procedure are provided in Appendix \ref{ss:experiments}.
First, Fig. \ref{fig:Pareto} (a) shows the training time and peak memory consumption when using the exact solution and when using DSTD ($M=10,~20$).
All results in this paper were obtained using a single GPU (NVIDIA GH200 120GB).
The horizontal axis of this graph, the channel width $N_\text{ch}$, represents the width of the convolutional network; a larger value corresponds to a larger number of input spikes to each LIF model.
The batch size was set to 50.
When the exact solution was used, both the computation time and peak memory consumption increased nonlinearly as the channel width $N_\text{ch}$ increased. 
On the other hand, when DSTD is used, the computation time and peak memory consumption showed an approximately linear trend, resulting in substantially more efficient training when $N_\text{ch}$ is large.

Figures \ref{fig:Pareto} (b) and (c) show the spike-time distributions of each layer in the 9-layer network trained on the CIFAR-10 dataset.
In each panel, the spike-time distributions are shown from the first layer at the top to the second layer and subsequent layers below, and the bottom panel corresponds to the output layer.
The spike-time distributions were obtained by aggregating over all test samples and all neurons within each layer.
The green masks indicate the time intervals during which neurons in each layer operate, and the width of each interval is $\tau_\text{width}$.
As described above, this operating interval is shifted by $\tau_\text{shift}$ each time the layer depth increases by one.
Input spikes are projected within the time interval $[0,1]$.
Figure \ref{fig:Pareto} (b) shows the results when the time-window width in each layer is $\tau_\text{width} \approx 1.36$, whereas Fig. \ref{fig:Pareto}(c) shows the results for $\tau_\text{width} = 1.00$.
In both cases, neurons in each layer receive spikes from the preceding layer and fire within the operating time-window width $\tau_\text{width}$.
These results indicate that the temporal penalty term introduced into each layer successfully encourages neurons to fire within the prescribed time window.

A smaller time-window width $\tau_\text{width}$ enables faster output generation; however, the time-window width $\tau_\text{width}$ and the classification accuracy exhibit a tradeoff.
To demonstrate this tradeoff, we performed multi-objective optimization with respect to accuracy and the interval width $\tau_\text{width}$ (see Appendix \ref{ss:Pareto} for details),
and the results are shown in Fig. \ref{fig:Pareto} (d).
Multi-objective optimization was performed for three different LIF models, $(\tau_v, \tau_I)=(\infty, \infty), (\infty, \tau), (2\tau, \tau)$, and for different network widths $N_\text{ch}$.
In all cases, we observed a general tradeoff in which the accuracy improved as the time-window width $\tau_\text{width}$ increased.
It should be noted that hyperparameter search is computationally intensive; therefore, the tradeoff curves obtained here should be regarded as empirical.

Furthermore, to examine the validity of training with DSTD, Fig. \ref{fig:Pareto} (e) shows the change in accuracy as the number of DSTD steps $M$ was varied.
Note that $M=40$ was used during testing.
For all LIF models and channel widths $N_\text{ch}$, approximately $M=10$ was sufficient to obtain high classification accuracy, and the accuracy was already saturated by around $M=20$. 
Training was unstable for the model with $(\tau_v, \tau_I)=(\infty, \tau)$ when a small number of DSTD steps was used ($M<15$), but became stable for $M\geq15$.

\begin{figure*}
\centering
\includegraphics[clip, width=0.95\textwidth]{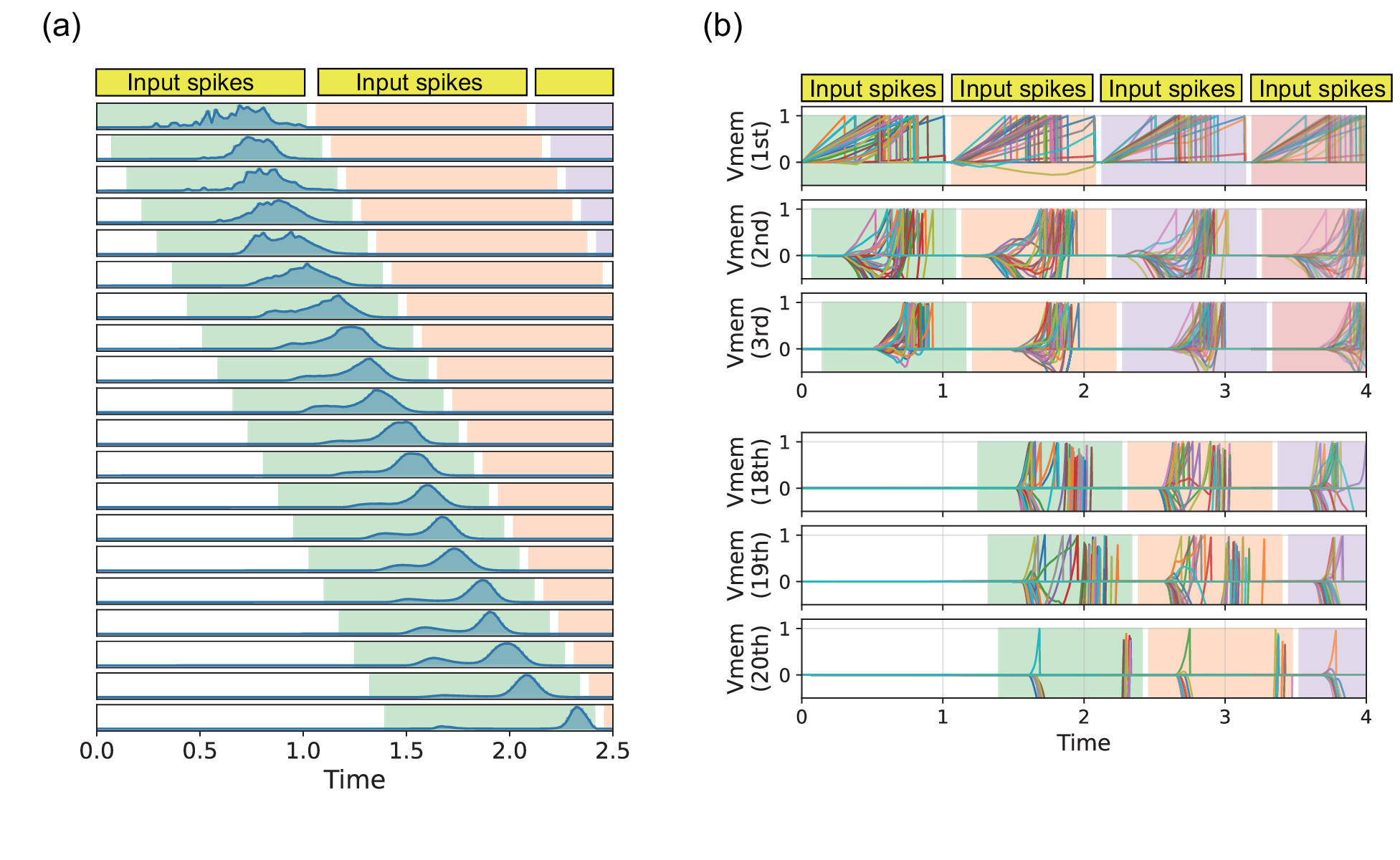}
\caption{\textbf{Training results for the 20-layer model (Syn-SNN-20).}
(a) Each panel shows the spike-time distribution of each layer, from the first layer at the top to the 20th layer at the bottom. The distributions were obtained by aggregating over all the test samples and all the neurons within each layer. Input spikes are projected in the interval $[0,1]$. The operating region of neurons in each layer, whose width is $\tau_\text{width}$, is masked in green.
(b) Examples of the temporal evolution of the membrane potentials in the first three layers and the last three layers. Pipeline operation is shown for the case in which four data samples are input sequentially. In this scenario, the next data sample is input after a margin time of 0.05 following the end of the processing time of the previous sample.
The background is masked with the same color for processing corresponding to the same data sample. 
}
\label{fig:ResNet}
\end{figure*}

Finally, we tested whether the proposed method can stably train a substantially deeper SNN.
We trained a 20-layer Syn-SNN (Syn-SNN-20) on Fashion-MNIST.
To improve training efficiency, we introduced a skip-and-delay mechanism corresponding to residual connections \cite{He2016deep} (see Appendix \ref{ss:architectures} for details).
The spike-time distribution of the trained model is shown in Fig. \ref{fig:ResNet} (a).
The time-window width was $\tau_\text{width}=1.01$.
Despite the depth of the network, spikes were propagated through the layers within the prescribed operating windows, indicating that the temporal penalty can organize spike propagation even in a 20-layer architecture.
Figure \ref{fig:ResNet} (b) further shows the temporal evolution of the membrane potentials in the first three layers and the last three layers.
To make the pipeline operation intuitive, we show the case in which four data samples were sequentially input with an inter-data margin of 0.05.
In this deep architecture, the benefit of pipeline processing becomes substantial: the first layer begins processing the $(k+1)$ th data sample before the 20th layer finishes processing the $k$ th data sample.
Thus, the proposed synfire-chain-like dynamics not only stabilizes the training of deep continuous-time SNNs but also enables pipeline processing of successive input samples.

\section{Discussion}

Over the past decade, the application of error backpropagation has made it possible to train multilayer SNNs with relatively high performance and substantially greater ease \cite{Dampfhoffer2023backpropagation}.
Surrogate-gradient learning is the most widely used approach for training deep SNNs.
In a typical implementation, SNNs are first discretized in time, spikes are regarded as binary activations, and their derivatives are replaced by either straight-through estimators \cite{Bengio2013estimating} or other smooth surrogate functions \cite{Zenke2018super,Neftci2017event,Wu2018spatio,Neftci2019surrogate}.
The theoretical validity of this approach remains under active discussion \cite{Arya2022automatic,Gygax2025elucidating}.
However, the gradients obtained by SGM are inherently heuristic, and it is fundamentally difficult for this approach to exploit the precise timing of spikes for information processing.

Continuous-time SNNs, in contrast, can realize ideal temporal coding because they allow precise control of spike timing and exact gradient computation.
This line of research was pioneered by Bohte et al., who derived a simple learning rule for computing exact gradients in continuous-time SNNs by using SNNs based on time-to-first-spike (TTFS) coding, in which the number of spikes per neuron is constrained to at most one \cite{Bohte2002error}.
Subsequently, de Montigny and Mâsse derived analytical expressions for firing times, thereby deriving computationally efficient backpropagation rules for multilayer TTFS-SNN architectures \cite{Montigny2016analytical}. 
Related analytical TTFS formulations were subsequently developed and applied to more modern network architectures and benchmark datasets \cite{Mostafa2018supervised,Comsa2021temporal,Goltz2021fast,Sakemi2023supervised,Zhang2021rectified}.
Learning rules for TTFS-SNNs have also been extended to promote even sparser firing \cite{Sakemi2023sparse} and to incorporate models with delay pathways \cite{Goltz2025DelGrad}.
In addition to these TTFS-based approaches, the constraint on the number of spikes can be removed by similarly using analytical solutions for multiple-spike regimes \cite{Yamamoto2024can,Bacho2023exploring,Morrill2026bullet}. 
Recently, quadratic integrate-and-fire models have been adopted to remove spike discontinuities \cite{Klos2025smooth,Wenig2026quadratic}.
Other methods have also been proposed, including approaches based on time discretization and approximate gradients \cite{Kheradpisheh2019s4nn}, gradient computation based on rate coding \cite{Kim2020unifying,Jin2018hybrid}, and extensions to stochastic processes \cite{Holberg2024exact}.
Nevertheless, training continuous-time models is generally computationally demanding, and in the case of SNNs in particular, the presence of firing dynamics has made the training of large-scale models extremely difficult (see Appendix \ref{ss:benchmarking} for recent architectures).

In this study, we showed that the efficient training of multilayer continuous-time SNNs can be achieved by extending the DSTD method, originally proposed as an efficient training method for linear time-varying systems \cite{Sakemi2025harnessing}, to LIF models.
Our experiments demonstrated up to a 20-fold speedup and approximately 100-fold lower peak memory usage.
Furthermore, we showed that synfire-chain dynamics can resolve both the dead neuron problem, a major source of training instability in multilayer SNNs, and the degradation of data-processing efficiency, or throughput, associated with network depth.
We also demonstrated that synfire-chain dynamics can be readily acquired through training simply by specifying the firing time window for each layer.

An alternative to the analytical-solution-based learning rules considered in this study is the direct optimization of ordinary differential equations, or more precisely hybrid systems, using adjoint methods.
The adjoint method is known as a general framework for error backpropagation \cite{Lecun1988theoretical}.
Its application to SNNs was reported by Kuroe et al. \cite{Kuroe2006learning,Kuroe2010learning,Selvaratnam2000learning}.
Huh et al. further proposed an adjoint-based learning rule for an SNN model with smoothed synaptic currents \cite{Huh2018gradient}.
Wunderlich et al., similarly to Kuroe et al., formulated learning as event-driven error backpropagation while exactly handling discontinuous firing processes \cite{Wunderlich2021event}.
Lee et al. formulated exact gradient computation in the form of forward differentiation \cite{Lee2023exact}.
In addition, event-driven exact-gradient methods capable of learning delays have been proposed \cite{Meszaros2025efficient}.
Although adjoint-based error backpropagation provides a general framework, it usually requires ordinary differential equations to be solved with fine time steps in both the forward and backward directions, resulting in poor computational efficiency.
Consequently, reported studies have been limited to small network architectures (see Appendix \ref{ss:benchmarking} for recent architectures).

The temporal coding considered in this study is also an important concept in neuroscience.
Although there are multiple views on how spikes code information in the brain \cite{Fujii1996dynamical}, two major perspectives are rate coding, which assumes that information is represented by spike frequency, and temporal coding, which assumes that information is also carried by spike timing \cite{Panzeri2010Sensory,Yuste2024Neuronal,Senkowski2024multi}.
Recent advances in the measurement of spike signals have provided several experimental results supporting temporal coding \cite{Abeles1993spatiotemporal,Mainen1995reliability,Lestienne2001spike,Gollisch2008rapid,Sotomayor2025firing,Xie2024neuronal,Zhu2025temporal}.
Temporal coding is important both for understanding the brain and for engineering applications because it enables low-energy information processing with few spikes and fast signal transmission \cite{Roy2019towards}.
In biological brains, temporal coding is thought to be used at least in part.
For example, indirect evidence has been reported in flies \cite{Lestienne2001spike}, salamanders \cite{Gollisch2008rapid}, and, more recently, the human brain \cite{Xie2024neuronal}.
Moreover, recent large-scale measurement experiments are beginning to reveal more reliable mechanisms of temporal coding \cite{Sotomayor2025firing}.
Synfire chains are also considered one possible form for realizing temporal coding \cite{Abeles1991corticonics, Diesmann1999stable,Ikegaya2004synfire,Zheng2014robust,Moldakarimov2015feedback}.
Because the continuous-time SNNs used in this study can learn spike timing precisely, they provide an ideal mathematical model for investigating the mechanisms of temporal coding.
Since DSTD now makes it possible to train large-scale models, future comparisons with neurophysiological experiments may become feasible.

Because SNNs process information using spikes, as in the brain, they are expected to enable ultra-low-power AI through hardware implementation \cite{Kudithipudi2025neuromorphic}.
In particular, analog hardware implementations are expected to achieve high power efficiency through analog in-memory computing \cite{Sakemi2026analog}.
TTFS coding is especially important for energy-efficient hardware operation because it can process information with a small number of spikes \cite{Sakemi2023supervised,Oh2022neuron}.
Continuous-time SNNs enable natural mapping onto analog systems that operate in continuous time.
This contrasts with discretized SNNs trained by SGM, for which mapping to continuous-time systems requires an additional conversion step.
In addition, many implementations of continuous-time SNNs have been reported \cite{Pehle2022brainscales2,Uenohara2022mixed,Richter2024dynap,Moriya2025analog,Duran2026cmos}.
The synfire-chain dynamics developed in this study is expected to provide a mechanism that combines high throughput with high power efficiency, because it allows data to be input sequentially even in deep multilayer networks.

As discussed above, the Syn-SNN model proposed in this study and the DSTD-based training method provide a useful approach across machine learning, neuroscience, and AI hardware.
However, as is generally the case for multilayer SNNs, the theoretical understanding of their training mechanisms remains insufficient.
For example, although the distribution of initial weights is an important parameter in deep learning models, the appropriate distribution for multilayer SNNs remains unclear, unlike in the case of ANNs \cite{Glorot2010understanding}.
In Syn-SNNs, it is also necessary to impose appropriate training conditions for realizing synfire-chain dynamics.
For these reasons, SNNs, including Syn-SNNs, require extensive searches over many hyperparameter sets to achieve high learning performance.
Developing more efficient search methods, or reducing the number of hyperparameters through theoretical understanding, will be important for training even larger-scale models.
Moreover, we showed that DSTD can train a large-scale SNN with small step sizes $M$; however, the theoretical understanding of this learning process with approximated gradient should be examined in future work.

\section*{Acknowledgments}
This work was partially supported by 
a project,
JPNP14004, commissioned by the New
Energy and Industrial Technology Development Organization
(NEDO), JSPS KAKENHI Grant Number 25K00148,
JST Research and Development Program for Next-generation
Edge AI Semiconductors, Grant Number JPMJES2511, 
Moonshot R\&D Grant Number JPMJMS2021, and
Cross-ministerial Strategic Innovation Promotion Program (SIP), the 3rd period of SIP, grant no. JPJ012207.

\section*{LLM Usage}
We used large language model-based tools to assist with language editing and code debugging. All outputs were reviewed and verified by the authors, who take full responsibility for the final manuscript, code, and results.

\newpage
\appendix

\section{Model} \label{ss:model}

\begin{table}[tb]
\centering
\caption{Relationship between the time constants $(\tau_v,\tau_I)$ and the spike kernel $K(t)$ in the analytical solution of $v(t)$. Here, $\theta(t)$ denotes the Heaviside step function.
}
\begin{tabular}{cc}
\hline
$(\tau_v,\tau_I)$ & $K(t)$ \\
\hline
$\tau_v\neq \tau_I$     & $\displaystyle\frac{\tau _v \tau_I}{\tau _v - \tau _I} \left[ e^{-\frac{t}{\tau_v}} - e^{-\frac{t}{\tau_I}} \right]\theta(t)$ \\
$(\tau,\tau)$     & $\displaystyle t\,e^{-\frac{t}{\tau}} \theta(t)$ \\
$(2\tau,\tau)$    & $\displaystyle 2\tau\left(e^{-\frac{t}{2\tau}} - e^{-\frac{t}{\tau}}\right) \theta (t)$ \\
$(\infty,\tau)$   & $\displaystyle \tau\left(1 - e^{-\frac{t}{\tau}}\right) \theta(t)$ \\
$(\infty,\infty)$ & $t \theta(t)$ \\
\hline
\end{tabular}
\label{tab:kernel}
\end{table}

In this study, we consider multilayer SNNs in which each neuron is described by a LIF model.
Under the constraint that each neuron fires at most once, namely time-to-first-spike (TTFS) coding, the LIF model is defined as follows:
\begin{align}
\begin{split}
\frac{dv_i^{(l)}}{dt} (t) &= -\frac{1}{\tau_v}v_i^{(l)}(t) + I_i^{(l)}(t), \\
    \frac{dI_i^{(l)}}{dt} (t) &= -\frac{1}{\tau_I}I_i^{(l)}(t) + \sum_{j=1}^{N^{(l-1)}}  w_{ij}^{(l)} \delta\left(t-t_j^{(l-1)}\right). 
\end{split} \label{eq:LIF_model_TTFS}
\end{align}
Here, $v_i^{(l)}(t)$ denotes the membrane potential of neuron $i$ in layer $l$, and $I_i^{(l)}(t)$ denotes the synaptic current of the same neuron.
The parameters $\tau_v$ and $\tau_I$ are the time constants of the membrane potential and synaptic current, respectively.
The variable $t_j^{(l-1)}$ represents the firing time of neuron $j$ in layer $l-1$.
The analytical solution of the membrane potential is given by
\begin{align}
 v_i^{(l)}(t) &= \sum_{j=1}^{N^{(l-1)}}  w_{ij}^{(l)} K\left( t - t_j^{(l-1)} \right).  \label{eq:Vmem_analytical}
\end{align}
Here, $K(\cdot)$ is the spike kernel function, whose form depends on the values of $\tau_v$ and $\tau_I$ (Table \ref{tab:kernel}).

\begin{table}[tb]
\centering
\caption{Expressions of the firing time candidate $t_{ik}^{(l)}$ corresponding to each pair of time constants $(\tau_v, \tau_I)$. For simplicity, the input spikes are assumed to be sorted, and the weights are sorted accordingly, such that $t_1 \le t_2 \le \cdots\le t_N$.}
\begin{tabular}{lccc}
\hline
$(\tau_v,\tau_I)$ &  $t_{ik}^{(l)}$ & $a_{ik}^{(l)}$ & $b_{ik}^{(l)}$ \\
\hline
$(\tau,\tau)$ 
&$\displaystyle
\tau\left[\frac{b_{ik}^{(l)}}{a_{ik}^{(l)}}
-W_0\!\Biggl(
  -\frac{V_{\mathrm{th}}/\tau}{a_{ik}^{(l)}}
   \,e^{\frac{b_{ik}^{(l)}}{a_{ik}^{(l)}}}
\Biggr)\right]
$
&  $\displaystyle \sum_{j=1}^k w_{ij}^{(l)}  \  e^\frac{t_j^{(l-1)}}{\tau}$ 
& $\displaystyle \sum_{j=1}^k w_{ij}^{(l)}\frac{t_j^{(l-1)}}{\tau} 
e^{\frac{t_j^{(l-1)}}\tau}$ \\[25pt]
$(2\tau,\tau)$ 
& $\displaystyle
2\tau \,\ln\!\left[
  \frac{2a_{ik}^{(l)}}{\,b_{ik}^{(l)}
           +\sqrt{(b_{ik}^{(l)})^2
                   -2\,a_{ik}^{(l)}{V_{\mathrm{th}}}/{\tau}}}
       
\right]
$ &  $\displaystyle \sum_{j=1}^k w_{ij}^{(l)}  \  e^\frac{t_j^{(l-1)}}{\tau}$ &$\displaystyle \sum_{j=1}^k w_{ij}^{(l)}  \  e^\frac{t_j^{(l-1)}}{2\tau}$\\[15pt]
$(\infty,\tau)$ 
&$\displaystyle
\tau \,\ln\!\left[
  \frac{a_{ik}^{(l)}}
       {b_{ik}^{(l)}
        -{V_{\mathrm{th}}}/{\tau}}
\right]$ 
&  $\displaystyle \sum_{j=1}^k w_{ij}^{(l)}  \  e^\frac{t_j^{(l-1)}}{\tau}$ 
& $\displaystyle \sum_{j=1}^k w_{ij}^{(l)}  $\\[15pt]
$(\infty,\infty)$ 
&$\displaystyle
\frac{V_{\mathrm{th}}+a_{ik}^{(l)}}
     {b_{ik}^{(l)}}$ 
& $\displaystyle\sum_{j=1}^k w_{ij}^{(l)}t_j^{(l-1)}  $ 
&  $ \displaystyle\sum_{j=1}^k w_{ij}^{(l)}  $
\\
\hline
\end{tabular}
\label{tab:spike_timing}
\end{table}

A LIF neuron fires and generates a spike when its membrane potential reaches the threshold $V_\text{th}$.
Thus, the firing time $t_i^{(l)}$ must satisfy
\begin{align}
v_i^{(l)}(t_i^{(l)}) = V_\text{th}. 
\end{align}
In principle, the firing time can be computed using the analytical solution.
However, in continuous-time LIF models, it is necessary to identify the index set of input spikes that arrive before the neuron fires, namely $\mathcal{S}_i^{(l)}=\{j|t_j^{(l-1)}\le t_i^{(l)} \}$.
This is typically done as follows \cite{Mostafa2018supervised,Zhou2021temporal}.

First, the indices $j$ of the neurons in layer $l-1$ are sorted in ascending order of their firing times:
\begin{align}
    \tilde{t}^{(l-1)}_1 \le \tilde{t}_2^{(l-1)} \le \cdots  \le \tilde{t}_{N^{(l-1)}}^{(l-1)}. 
\end{align}
The corresponding weights, sorted in the same order, are denoted by
 $\tilde{w}_{ij}^{(l)}$. 
Then, within the time interval $[\tilde{t}_k^{(l-1)}, \tilde{t}_{k+1}^{(l-1)}]$, the membrane potential is given by
\begin{align}
 v_i^{(l)}(t) &= \sum_{j=1}^{k}  \tilde{w}_
 {ij}^{(l)} K\left( t - \tilde{t}_j^{(l-1)} \right). 
\end{align}
We define the firing time within this interval as $t_{ik}^{(l)}$, whose explicit expressions are given in Table  \ref{tab:spike_timing}. 
Finally, the true firing time is obtained as
\begin{align}
    t_i^{(l)}=\min_k \{t_{ik}^{(l)}| \tilde{t}_k^{(l-1)} \le t_{ik}^{(l)} < \tilde{t}_{k+1}^{(l-1)}\},  \label{eq:timing_exact}
\end{align}
where $\tilde{t}_{N^{(l-1)}+1}^{(l-1)}=\infty$. 
Note that $t_{ik}^{(l)}$ can be computed in parallel for different $k$ using a prefix scan \cite{Zhou2021temporal,Yamamoto2024can,Sakemi2023sparse,Morrill2026bullet,Che2026parallel}.
Algorithm \ref{alg:TTFS} shows the pseudocode of the TTFS-SNN. 
The learning properties of multilayer SNNs with TTFS coding have been investigated for $(\tau_v, \tau_I) = (\infty, \infty)$ \cite{Sakemi2023supervised,Zhang2021rectified,Sakemi2023sparse}, $(\tau_v, \tau_I) = (\infty, \tau)$~\cite{Mostafa2018supervised}, $(\tau_v, \tau_I) = (2\tau, \tau)$~\cite{Goltz2021fast}, and $(\tau_v, \tau_I) = (\tau, \tau)$~\cite{Montigny2016analytical,Comsa2021temporal}.

 \begin{algorithm}[tb]
\caption{TTFS-SNN}
\label{alg:TTFS}
\begin{algorithmic}[1] 
\State Obtain input spikes $t_i^{(0)}$ from data  
\For{all layers ($l=1,...,L$)}
  \State Obtain $\tilde{t}_j^{(l-1)}$ by sorting $t_j^{(l-1)}$
  \State Calculate $t_{ik}^{(l)}$ from Table \ref{tab:spike_timing}
  \State Obtain $t_i^{(l)}$ with Eq. \eqref{eq:timing_exact}
\EndFor
\end{algorithmic}
\end{algorithm}

\section{Differentiable spike-time discretization for LIF models} \label{ss:DSTD}

The spike-time computation shown in Table \ref{tab:spike_timing} and Eq. \eqref{eq:timing_exact} requires the evaluation of as many candidate spike times as the number of input spikes.
As a result, the memory requirement of each layer becomes $\mathcal{O}\left(N^{(l)}N^{(l-1)}\right)$.
Because of this computational cost, large-scale continuous-time TTFS-SNNs have remained relatively limited in size, except for previous work that relied on GPU clusters \cite{Zhou2021temporal}.
Differentiable spike-time discretization (DSTD) addresses this problem.

\subsection{Differentiable spike-time discretization}

Differentiable spike-time discretization (DSTD) \cite{Sakemi2025harnessing} is a method for improving computational efficiency by differentiably approximating input spike trains as spike trains on predefined discrete time points.
We first define the discrete time points in each layer as follows:
\begin{flalign}
&T^{(l)}_m =  T_\text{first}^{(l)} + m \Delta t - t_\text{offset}^{(l)},~(m=0,1,\dots, M),  \\
&-\frac{\Delta t}{2} < t_\text{offset}^{(l)} \le \frac{\Delta t}{2} 
\end{flalign}

\begin{table}[tb]
\centering
\caption{Candidate spike times $t_{iq}^{(l)}$ obtained with DSTD for different time-constant settings $(\tau_v, \tau_I)$.}
\begin{tabular}{lcccc}
\hline
$(\tau_v,\tau_I)$ &  $t_{iq}^{(l)}$ & $a_{iq}^{(l)}$ & $b_{iq}^{(l)}$ \\
\hline
$(\tau,\tau)$ 
&$\displaystyle
\tau\left[\frac{b_{iq}^{(l)}}{a_{iq}^{(l)}}
-W_0\!\Biggl(
  -\frac{V_{\mathrm{th}}/\tau}{a_{iq}^{(l)}}
   \,e^{\frac{b_{iq}^{(l)}}{a_{iq}^{(l)}}}
\Biggr)\right]
$
&  $\displaystyle\sum_{m=0}^q \hat{w}_{im}^{(l)}  \  e^\frac{T_m^{(l)}}{\tau}$  
& $\displaystyle\sum_{m=0}^q \hat{w}_{im}^{(l)}\frac{T_m^{(l)}}{\tau} 
e^{\frac{T_m^{(l)}}\tau}$ \\[25pt]
$(2\tau,\tau)$ 
&$\displaystyle
2\tau \,\ln\!\left[
  \frac{\,b_{iq}^{(l)}
           -\sqrt{(b_{iq}^{(l)})^2
                   -2\,a_{iq}^{(l)}{V_{\mathrm{th}}}/{\tau}}}
       {V_{\mathrm{th}}/\tau}
\right]
$ 
&  $\displaystyle\sum_{m=0}^q \hat{w}_{im}^{(l)}  \  e^\frac{T_m^{(l)}}{\tau}$ 
&$\displaystyle\sum_{m=0}^q \hat{w}_{im}^{(l)}  \  e^\frac{T_m^{(l)}}{2\tau}$ \\[15pt]
$(\infty,\tau)$ 
&$\displaystyle
\tau \,\ln\!\left[
  \frac{a_{iq}^{(l)}}
       {b_{iq}^{(l)}
        -{V_{\mathrm{th}}}/{\tau}}
\right]$ 
&  $\displaystyle\sum_{m=0}^q \hat{w}_{im}^{(l)}  \  e^\frac{T_m^{(l)}}{\tau}$ &  $ \sum_{m=0}^q \hat{w}_{im}^{(l)}  $\\[15pt]
$(\infty,\infty)$ 
&$\displaystyle
\frac{V_{\mathrm{th}}+a_{iq}^{(l)}}
     {b_{iq}^{(l)}}$ 
& $\displaystyle\sum_{m=0}^q \hat{w}_{im}^{(l)}T_m^{(l)}  $ 
&  $\displaystyle \sum_{m=0}^q \hat{w}_{im}^{(l)}  $
\\
\hline
\end{tabular}
\label{tab:timing_DSTD}
\end{table}

DSTD transforms a spike at $t_j^{(l-1)}$ into a real-valued spike train on these discrete time points: 
\begin{align}
    \sum_{m=0}^M s_{jm}^{(l-1)}\delta\left(t-T_m^{(l)}\right) = \text{DSTD}\left( \delta \left( t - t_j^{(l-1)}\right)\right). \label{eq:DSTD_element}
\end{align}
Here, $s_{jm}^{(l-1)} \in \mathbb{R}$ represents the magnitude of the spike at the discrete time point $T_m^{(l)}$.
When all input spike trains are transformed by DSTD, the approximated membrane potential $\hat{v}_i^{(l)}(t)$ in the time interval $[T_q^{(l)}, T_{q+1}^{(l)}]$ is given by
\begin{align}
    \hat{v}_i^{(l)}(t) & =  \sum_{j=1}^{N^{(l-1)}}  w_{ij}^{(l)} \sum_{m=0}^{q}  s_{jm}^{(l-1)} K\left(t-T_m^{(l)}\right) \\
    &= \sum_{m=0}^{q} \left(\sum_{j=1}^{N^{(l-1)}}  w_{ij}^{(l)} s_{jm}^{(l-1)}\right)  K\left(t-T_m^{(l)}\right) \\
    &= \sum_{m=0}^{q} \hat{w}_{im}^{(l)} K\left(t-T_m^{(l)}\right).
\end{align}
Here, we define
\begin{align}
\hat{w}_{im}^{(l)}:=\sum_{j=1}^{N^{(l-1)}}  w_{ij}^{(l)} s_{jm}^{(l-1)}.
\end{align}
This quantity can be interpreted as the total weight of all spikes arriving at time $T_m^{(l)}$.
The candidate spike time $t_{iq}^{(l)}$ obtained from the firing condition of the membrane potential is derived by solving $\hat{v}_i^{(l)}(t_{iq}^{(l)})=V_\text{th}$, as shown in Table \ref{tab:timing_DSTD}.
Using these candidate spike times, the spike time under DSTD is obtained as
\begin{align}
    \hat{t}_i^{(l)} = \min_q \{t_{iq}^{(l)}|T_q^{(l)} \le t_{iq}^{(l)} < T_{q+1}^{(l)}  \}. \label{eq:timing_DSTD}
\end{align}
DSTD reduces the number of candidate spike times from
$N^{(l-1)}$, corresponding to the presynaptic spike times in the exact
formulation, to the fixed number of DSTD intervals, $M$.
The coefficients for the candidate intervals are evaluated using parallel
prefix-scan operations in both formulations.
Thus, DSTD preserves the same basic candidate-evaluation procedure while
reducing the length of the candidate dimension from $N^{(l-1)}$ to $M$.
In the next subsection, we describe the concrete transformation rule, that is, the values of $s_{jm}^{(l-1)}$.

\subsection{Derivation of DSTD kernels for LIF models}

Discretizing input spikes using DSTD reduces the memory requirement, but the membrane potential obtained in this way is an approximation.
To make this approximation as accurate as possible, we derive the optimal transformation according to the time constants of the LIF model, $(\tau_v, \tau_I)$.
In this paper, we define the optimal approximation as one that satisfies
\begin{align}
    \hat{v}_i^{(l)}(T_m) = v_i^{(l)}\left( T_m \right) \text{ for } m=0,1,\dots,M.
\end{align}
We then consider a transformation
\begin{align}
    (s_{j0}, s_{j1}, \dots , s_{jM} ) = \text{DSTD}(t_j)
\end{align}
that satisfies this condition.
Since the membrane potential of the LIF model, Eq. \ref{eq:Vmem_analytical}, is expressed as a sum over input spikes, it is sufficient to consider the case of a single input spike ($N=1$).
For notational simplicity, we omit the layer index $l$ and the neuron index $i$.

Let $T_{m-1} < t_1 < T_m$. In this case,
\begin{align}
    v(t) &= w_1 K(t - t_1) \\
    &= w_1 \frac{\tau _v \tau_I}{\tau _v - \tau _I} \left[ e^{-\frac{t-t_1}{\tau_v}} - e^{-\frac{t-t_1}{\tau_I}} \right] \\
    &= w_1 \frac{\tau _v \tau_I}{\tau _v - \tau _I} \left[ e^{-\frac{T_m-t_1}{\tau_v}} e^{-\frac{t-T_m}{\tau_v}} - e^{-\frac{T_m - t_1}{\tau_I}}  e^{-\frac{t-T_m}{\tau_I}} \right],
\end{align}
and
\begin{align}
    \hat{v}(t) &=   w_{1} \left[ s_{1~\!m-1}  K\left(t-T_{m-1}\right) + s_{1~\!m}  K\left(t-T_{m}\right) \right] \\
    &=  \frac{w_{1}\tau _v \tau_I}{\tau _v - \tau _I}
    \left[ s_{1~\!m-1}  \left(e^{-\frac{t-T_{m-1}}{\tau_v}} - e^{-\frac{t - T_{m-1}}{\tau_I}}\right) 
    + s_{1~\!m}  \left(e^{-\frac{t-T_m}{\tau_v}} - e^{-\frac{t-T_m}{\tau_I}} \right)\right] \\
    &=  \frac{w_{1}\tau _v \tau_I}{\tau _v - \tau _I}
    \left[ \left(e^{-\frac{T_m - T_{m-1}}{\tau_v}} s_{1~\!m-1} + s_{1m} \right) e^{-\frac{t-T_m}{\tau_v}} 
    -   \left(e^{-\frac{T_m - T_{m-1}}{\tau_I}} s_{1~\!m-1} + s_{1~\!m}\right)e^{-\frac{t-T_m}{\tau_I}} \right]
\end{align}

For $\hat{v}(t) = v(t)$ to hold for $t \ge T_m$, it is sufficient that the following equation is satisfied: 
\begin{align}
    \begin{pmatrix}
        e^{-\frac{T_m - T_{m-1}}{\tau_v}} & 1 \\
        e^{-\frac{T_m - T_{m-1}}{\tau_I}} & 1 
    \end{pmatrix}
    \begin{pmatrix}
        s_{1~m-1} \\ s_{1m}
    \end{pmatrix} = 
    \begin{pmatrix}
        e^{-\frac{T_m-t_1}{\tau_v}} \\  e^{-\frac{T_m - t_1}{\tau_I}}
    \end{pmatrix}.
\end{align}
When $\tau_v \neq \tau_I$, the inverse matrix exists, yielding
\begin{align} 
s_{1~m-1} &=  \frac{e^{-\frac{T_m-t_1}{\tau_v}} - e^{-\frac{T_m - t_1}{\tau_I}}}{e^{-\frac{T_m - T_{m-1}}{\tau_v}} - e^{-\frac{T_m - T_{m-1}}{\tau_I}}} \\
s_{1m} &=  \frac{-e^{-\frac{T_m - T_{m-1}}{\tau_I}} e^{-\frac{T_m-t_1}{\tau_v}} + e^{-\frac{T_m - T_{m-1}}{\tau_v}} e^{-\frac{T_m - t_1}{\tau_I}}}{e^{-\frac{T_m - T_{m-1}}{\tau_v}} - e^{-\frac{T_m - T_{m-1}}{\tau_I}}}. 
\end{align}
By setting $T_{m}-T_{m-1}=\Delta t  ~(m=1,\dots,M)$, this can be rewritten as
\begin{align}
    s_{jm}=\begin{cases}
        \frac{-e^{-\frac{\Delta t}{\tau_I}} e^{-\frac{T_m - t_j}{\tau_v}} + e^{-\frac{\Delta t}{\tau_v}} e^{-\frac{ T_m -  t_j}{\tau_I}}}{e^{-\frac{\Delta t}{\tau_v}} - e^{-\frac{\Delta t}{\tau_I}}},&\text{ when }   T_{m-1} <t_j \le T_m \\
        \frac{e^{-\frac{T_m + \Delta t - t_j}{\tau_v}} - e^{\frac{T_m +\Delta t-  t_j}{\tau_I}}}{e^{-\frac{\Delta t}{\tau_v}} - e^{-\frac{\Delta t}{\tau_I}}}, &\text{ when }   T_{m} <t_j \le T_{m+1} \\
        0,&\text{ otherwise,}
    \end{cases}
\end{align}

When the membrane and synaptic time constants are equal, 
for $t\geq T_m$, the exact membrane potential generated by this spike is
\begin{align}
v(t)&=w_1(t-t_1)e^{-\frac{t-t_1}{\tau}} \\
&= w_1 \left[ e^{\frac{t_1}{\tau}}  \cdot te^{-\frac{t}{\tau}} - t_1 e^{\frac{t_1}{\tau}}  \cdot e^{-\frac{t}{\tau}}\right]
\end{align}
We approximate the original spike by two weighted spikes at
$T_{m-1}$ and $T_m$:
\begin{align}
\hat{v}(t)
&=
w_1
\left[
s_{1~m-1}K(t-T_{m-1})
+
s_{1m}K(t-T_m)
\right] \\
&= w_1 \left[ s_{1~m-1}(t-T_{m-1})e^{-\frac{t-T_{m-1}}{\tau}} + s_{1m} (t-T_{m})e^{-\frac{t-T_{m}}{\tau}} \right] \\
&= w_1 \left[ \left( s_{1~m-1} e^{\frac{T_{m-1}}{\tau}} + s_{1m} e^{\frac{T_m}{\tau}} \right) \cdot t e^{-\frac{t}{\tau}} - \left(s_{1~m-1} T_{m-1} e^{\frac{T_{m-1}}{\tau}} + s_{1m} T_m e^{\frac{T_m}{\tau}} \right)\cdot e^{-\frac{t}{\tau}} \right]. 
\end{align}
For $\hat{v}(t) = v(t)$ to hold for $t \ge T_m$, it is sufficient that the following equation is satisfied: 
\begin{align}
    s_{1~m-1} e^{\frac{T_{m-1}}{\tau}} + s_{1m} e^{\frac{T_m}{\tau}} &=  e^{\frac{t_1}{\tau}} \\
    s_{1~m-1} T_{m-1} e^{\frac{T_{m-1}}{\tau}} + s_{1m} T_m e^{\frac{T_m}{\tau}} &= t_1  e^{\frac{t_1}{\tau}}
\end{align}

\begin{align}
    \begin{pmatrix}
       e^{\frac{T_{m-1}}{\tau}} & e^{\frac{T_m}{\tau}} \\
         T_{m-1} e^{\frac{T_{m-1}}{\tau}} &  T_m e^{\frac{T_m}{\tau}} 
    \end{pmatrix}
    \begin{pmatrix}
        s_{1~m-1} \\ s_{1m}
    \end{pmatrix} = 
    \begin{pmatrix}
        e^{\frac{t_1}{\tau}} \\   t_1 e^{\frac{t_1}{\tau}}
    \end{pmatrix}.
\end{align}
Solving these equations gives 
\begin{align}
s_{1,m-1}
&=\frac{T_m-t_1}{\Delta t}
e^{\frac{t_1-T_{m-1}}{\tau}},
\label{eq:equal_tau_left_weight}\\
s_{1m}
&=\frac{t_1-T_{m-1}}{\Delta t}
e^{\frac{t_1-T_m}{\tau}}.
\label{eq:equal_tau_right_weight}
\end{align}
This can be rewritten as
\begin{align}
s_{jm}
= \begin{cases}
\frac{t_j-T_{m-1}}{\Delta t}
e^{\frac{t_j-T_m}{\tau}}, &\text{ when }   T_{m-1} <t_j \le T_m \\
\frac{T_{m+1}-t_j}{\Delta t}
e^{\frac{t_j-T_m}{\tau}}, &\text{ when }   T_{m} <t_j \le T_{m+1}\\
        0, &\text{ otherwise.}
\end{cases}
\end{align}

Solutions can also be obtained in the same manner for cases in which the time constants become infinite.

When $\tau_v = \infty$ and $\tau_I = \tau~ (<\infty)$:
 \begin{align}
s_{1~m-1} &= \frac{1 - e^{-\frac{T_m - t_1}{\tau}}}{1 - e^{-\frac{T_m - T_{m-1}}{\tau}}} \\
s_{1 m} &=  \frac{-e^{-\frac{T_m - T_{m-1}}{\tau}} 
+  e^{-\frac{T_m - t_1}{\tau}}}{1 - e^{-\frac{T_m - T_{m-1}}{\tau}}} 
\end{align}

\begin{align}
    s_{jm} &= \begin{cases}
        \frac{-e^{-\frac{\Delta t}{\tau_I}} 
+  e^{-\frac{T_m - t_j}{\tau_I}}}{1 - e^{-\frac{\Delta t}{\tau_I}}}  &\text{ when }   T_{m-1} <t_j \le T_m, \\
\frac{1 - e^{-\frac{T_m - t_j + \Delta t }{\tau_I}}}{1 - e^{-\frac{\Delta t}{\tau_I}}} &\text{ when }   T_{m} <t_j \le T_{m+1}, \\
        0, &\text{ otherwise,}
    \end{cases} 
\end{align}

When $\tau_v = \tau_I=\infty$:
\begin{align}
s_{1~m-1} &= \frac{T_m - t_1}{T_m - T_{m-1}} \\
s_{1 m} &= \frac{t_1 - T_{m-1}}{T_m - T_{m-1}}
\end{align}

\begin{align}
    s_{jm} &= \begin{cases}
        \frac{t_j - T_m + \Delta t}{\Delta t} &\text{ when }   T_{m-1} <t_j \le T_m, \\
        \frac{T_m + \Delta t - t_j}{\Delta t} &\text{ when }   T_{m} <t_j \le T_{m+1}, \\
        0, &\text{ otherwise,}
    \end{cases} \\
    &= \max \left( 1 - \frac{|T_m - t_j|}{\Delta t}, 0 \right)
\end{align}

\begin{table}[tb]
\centering
\caption{DSTD kernels. We assume $T_{m+1}-T_m = \Delta t$. }
\begin{tabular}{lcc}
\hline
$(\tau_v,\tau_I)$ & $K_\text{DSTD}(t_{<0})$ & $K_\text{DSTD}(t_{\ge0})$   \\
\hline
$\tau_v \ne \tau_I$ &  $\displaystyle \frac{-e^{-\frac{\Delta t}{\tau_I}} e^{\frac{t}{\tau_v}} + e^{-\frac{\Delta t}{\tau_v}} e^{\frac{  t}{\tau_I}}}{e^{-\frac{\Delta t}{\tau_v}} - e^{-\frac{\Delta t}{\tau_I}}}$ 
& $\displaystyle \frac{e^{\frac{t - \Delta t}{\tau_v}} - e^{\frac{ t -\Delta t}{\tau_I}}}{e^{-\frac{\Delta t}{\tau_v}} - e^{-\frac{\Delta t}{\tau_I}}}$
\\ 
$(\tau, \tau)$ 
&$\displaystyle \frac{t+\Delta t}{\Delta t}e^{t/\tau}$
&$\displaystyle \frac{\Delta t-t}{\Delta t}e^{t/\tau}$\\
$(2\tau, \tau)$ 
& $\displaystyle \frac{-e^{-\frac{\Delta t}{\tau}} e^{\frac{t}{2\tau}} + e^{-\frac{\Delta t}{2\tau}} e^{\frac{  t}{\tau}}}{e^{-\frac{\Delta t}{2\tau}} - e^{-\frac{\Delta t}{\tau}}}$ 
& $\displaystyle \frac{e^{\frac{t - \Delta t}{2\tau}} - e^{\frac{ t -\Delta t}{\tau}}}{e^{-\frac{\Delta t}{2\tau}} - e^{-\frac{\Delta t}{\tau}}}$\\
$(\infty, \tau)$ & $\displaystyle \frac{-e^{-\frac{\Delta t}{\tau}} 
+  e^{\frac{t}{\tau}}}{1 - e^{-\frac{\Delta t}{\tau}}}$
& $\displaystyle \frac{1 - e^{\frac{t - \Delta t}{\tau}}}{1 - e^{-\frac{\Delta t}{\tau}}}$ \\
$(\infty, \infty)$ & $\displaystyle \frac{t + \Delta t}{\Delta t} $
& $\displaystyle \frac{\Delta t- t}{\Delta t}$
\\\hline
\end{tabular}
\label{tab:DSTD_kernels}
\end{table}

By expressing these transformations using the DSTD kernel function $K_\text{DSTD}$ as 
\begin{align}
    s_{jm} = \begin{cases}
        K_\text{DSTD} \left( t_j - T_m \right), &\text{ when } T_{m-1}\le  t_j <T_{m+1}, \\
        0 &\text{ otherwise.}
        \end{cases} \label{eq:DSTD_kernel}
\end{align}
they can be summarized as shown in Table \ref{tab:DSTD_kernels}.
Note that the previous study \cite{Sakemi2025harnessing} used the DSTD function corresponding to the case $\tau_v=\tau_I=\infty$.

\begin{figure*}
\centering
\includegraphics[clip, width=\textwidth]{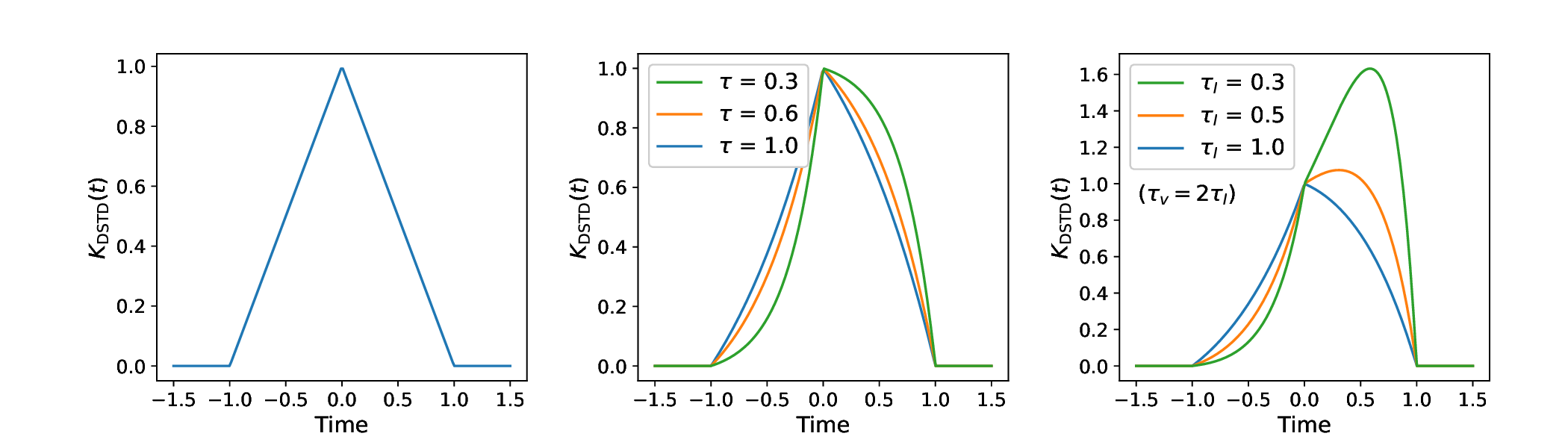}
\caption{{\bf DSTD kernels. }
DSTD kernel functions $K_\text{DSTD}(t)$ for various choices of time constants when $\Delta=1$.
(Left) $\tau_v = \tau_I = \infty$, (middle) $\tau_v=\infty, \tau_I = \tau$, and (right) $\tau_v = 2 \tau_I$.
}
\label{fig:DSTD_kernel}
\end{figure*}

Figure \ref{fig:DSTD_kernel} shows the DSTD kernel $K_\text{DSTD}$ for various choices of time constants in the LIF model.
The pseudocode of the TTFS-SNN using DSTD is shown in Algorithm \ref{alg:TTFS_DSTD}.

\begin{algorithm}[H]
\caption{TTFS-SNN with DSTD}
\label{alg:TTFS_DSTD}
\begin{algorithmic}[1] 
\State Obtain input spikes $t_i^{(0)}$ from data  
\State Let $\hat{t}_i^{(0)}=t_i^{(0)}$
\For{all layers ($l=1,...,L$)}
  \State Draw $t_\text{offset}^{(l)}$ from $(-\Delta t/2, \Delta t/2]$
  \State Calculate $s_{jm}^{(l-1)}$ with $\hat{t}_j^{(l-1)}$ using Eq. (\ref{eq:DSTD_kernel}).
  \State Calculate ${t}_{im}^{(l)}$ from Table \ref{tab:timing_DSTD}.
  \State Calculate $\hat{t}_i^{(l)}$ solving Eq. (\ref{eq:timing_DSTD}). 
\EndFor
\end{algorithmic}
\end{algorithm}

\subsection{Computational complexity} \label{ss:complexity}

Let
\[
N_{\mathrm{in}} = N^{(l-1)},
\qquad
N_{\mathrm{out}} = N^{(l)}.
\]
For simplicity, we omit the batch dimension.

Table~\ref{tab:computational_complexity} summarizes the computational
and memory complexities of the exact spike-time formulation and DSTD.
Here, the complexities are expressed in terms of total work.
The parallel depth of the prefix-scan operations is shown separately.

\begin{table}[t]
\centering
\caption{
Computational and memory complexities of the exact spike-time
formulation and DSTD for a single layer.
The activation-memory complexity denotes the memory required to retain
candidate-dependent intermediate variables for reverse-mode automatic
differentiation.
}
\label{tab:computational_complexity}
\begin{tabular}{lcc}
\hline
 & Exact formulation & DSTD \\
\hline
Candidate dimension
&
$N_{\mathrm{in}}$
&
$M$
\\

Spike-time sorting
&
$O(N_{\mathrm{in}}\log N_{\mathrm{in}})$
&
Not required
\\

Weight permutation
&
$O(N_{\mathrm{out}}N_{\mathrm{in}})$
&
Not required
\\

Input discretization
&
Not required
&
$O(N_{\mathrm{out}}N_{\mathrm{in}})$
\\

Candidate-coefficient evaluation
&
$O(N_{\mathrm{out}}N_{\mathrm{in}})$
&
$O(N_{\mathrm{out}}M)$
\\

Total computational work
&
$
O\!\left(
N_{\mathrm{in}}\log N_{\mathrm{in}}
+
N_{\mathrm{out}}N_{\mathrm{in}}
\right)
$
&
$
O\!\left(
N_{\mathrm{out}}
\left(
N_{\mathrm{in}}+M
\right)
\right)
$
\\

Parallel scan depth
&
$O(\log N_{\mathrm{in}})$
&
$O(\log M)$
\\

Candidate-related activation memory
&
$O(N_{\mathrm{out}}N_{\mathrm{in}})$
&
$O(N_{\mathrm{out}}M)$
\\
\hline
\end{tabular}
\end{table}

In the exact formulation, the presynaptic spike times are first sorted,
which requires
\[
O\!\left(N_{\mathrm{in}}\log N_{\mathrm{in}}\right)
\]
operations.
Because the ordering of the presynaptic spike times is common to all
postsynaptic neurons in the same layer, sorting is performed only once
for each input sample.
The same permutation is then applied to the synaptic weights, requiring
\[
O\!\left(N_{\mathrm{out}}N_{\mathrm{in}}\right)
\]
operations.

For each postsynaptic neuron, the cumulative coefficients required to
evaluate the candidate firing times can be calculated using parallel
prefix-scan operations.
The total work required to calculate the coefficients and evaluate all
candidate firing times is therefore
\[
O\!\left(N_{\mathrm{out}}N_{\mathrm{in}}\right),
\]
with a parallel depth of
\[
O\!\left(\log N_{\mathrm{in}}\right).
\]
Consequently, the total computational complexity of the exact
formulation is
\[
O\!\left(
N_{\mathrm{in}}\log N_{\mathrm{in}}
+
N_{\mathrm{out}}N_{\mathrm{in}}
\right).
\]

Although the total computational work is linear in the number of
presynaptic connections, the exact formulation generates
$N_{\mathrm{in}}$ candidate firing times for every postsynaptic neuron.
When these candidate-dependent intermediate variables are retained for
reverse-mode automatic differentiation, the corresponding
activation-memory requirement is
\[
O\!\left(N_{\mathrm{out}}N_{\mathrm{in}}\right).
\]

With DSTD, each presynaptic spike is mapped to at most two adjacent
discrete time points because the DSTD kernel has compact support.
By exploiting this sparsity, the discretized weighted inputs
$\hat{w}^{(l)}_{im}$ can be accumulated in
\[
O\!\left(N_{\mathrm{out}}N_{\mathrm{in}}\right)
\]
operations.
The cumulative coefficients and candidate firing times are subsequently
evaluated over the $M$ DSTD intervals, requiring
\[
O\!\left(N_{\mathrm{out}}M\right)
\]
operations, with a parallel prefix-scan depth of
\[
O(\log M).
\]
The total computational complexity of DSTD is therefore
\[
O\!\left(
N_{\mathrm{out}}
\left(
N_{\mathrm{in}}+M
\right)
\right),
\]
and its candidate-related activation-memory requirement is
\[
O\!\left(N_{\mathrm{out}}M\right).
\]

Thus, DSTD does not eliminate the linear cost associated with the
synaptic connections.
Instead, it replaces the input-dependent candidate dimension
$N_{\mathrm{in}}$ with the fixed dimension $M$.
When
\[
M \ll N_{\mathrm{in}},
\]
this substantially reduces the memory required to retain
candidate-dependent intermediate variables.
It also removes spike-time sorting and converts the candidate evaluation
into regular fixed-size tensor operations, which are well suited to
parallel execution on GPUs.
The resulting improvements in wall-clock time and peak memory
consumption are evaluated empirically in Figs.~\ref{fig:DSTD_exp} and~\ref{fig:Pareto}.

\subsection{Extension to multi-spike regime} \label{ss:multi-spikes}

When each neuron can fire multiple spikes, the LIF model is described as follows:
\begin{align}
    \frac{d v_i^{(l)}}{dt}(t) &= -\frac{1}{\tau_v} v_i^{(l)}(t) + I_i^{(l)}(t) \\
    \frac{d I_i^{(l)}}{dt}(t) &= -\frac{1}{\tau_I} I_i^{(l)}(t) + \sum_{j=1}^{N^{(l-1)}} \sum_{f=1}^F w_{ij}^{(l)} \delta(t-t_{j}^{(l-1),f}).
\end{align}
Here, $\tau_{v(I)}$ denotes the time constant of the membrane potential (synaptic current), $t_{j}^{(l-1), f}$ is the timing of the $f$-th spike arriving from neuron $j$ in layer $l-1$, $w_{ij}^{(l)}$ is the synaptic weight from neuron $j$ to neuron $i$, and $\delta(\cdot)$ is the Dirac delta function.
When the membrane potential reaches the firing threshold $V_\text{th}$, the neuron fires and generates a spike.
The membrane potential is then reset to $0$.
In contrast, the synaptic current is assumed not to be reset at the firing time \cite{Yamamoto2024can}.

After this neuron emits its $k$-th spike at time $t_{i}^{(l),k}$, the membrane potential can be written analytically as
\begin{align}
     v_i^{(l)}(t) &= \left(\sum_{j=1}^{N^{(l-1)}} \sum_{f=1}^F w_{ij}^{(l)} e^{-\frac{t_i^{(l),k} - t_j^{(l-1),f}}{\tau_I}}\theta(t_{i}^{(l),k} - t_j^{(l-1),f})\right) K(t-t_i^{(l),k}) 
     \nonumber\\ 
     &~~~~~~+  \sum_{j=1}^{N^{(l-1)}} \sum_{f=1}^F  w_{ij}^{(l)} \theta(t_j^{(l-1),f}-t_i^{(l),k}) \theta\left(t - t_j^{(l-1),f}\right) K\left( t - t_j^{(l-1),f} \right). \label{eq:Vmem_multiSpikes} 
\end{align}
Intuitively, the first term represents the contribution of all spikes that arrived before the $k$-th firing event, whereas the second term represents the contribution of the spikes that arrived after the $k$-th firing event.

Equation \eqref{eq:Vmem_multiSpikes} shows that the effect of each input spike is independently mediated through the corresponding kernel function.
Therefore, DSTD can be applied straightforwardly as follows.
First, DSTD is applied independently to each spike:
\begin{align}
    s_{jm}^{(l-1),f} &:= \begin{cases}
        K_\text{DSTD} \left( t_j^{(l-1),f} - T_m^{(l)} \right), &\text{ when } T_{m-1}^{(l)}\le  t_j^{(l-1),f} <T_{m+1}^{(l)}, \\
        0 &\text{ otherwise.}
        \end{cases} \\
s_{jm}^{(l-1)} &:= \sum_{f=1}^F s_{jm}^{(l-1),f}. 
\end{align}
We then consider the computation of the $(k+1)$-th firing time after the $k$-th firing event occurs at time $t_{ik}^{(l)}$.
Under the assumption that the synaptic current is not reset at the firing time, the membrane potential during the time interval $[T_m, T_{m+1}]$ after firing can be computed as
\begin{align}
     v_i^{(l)}(t) &= \left[\sum_{j=1}^{N^{(l-1)}} w_{ij}^{(l)} \sum_{q=0}^m \underbrace{\left(\sum_{f=1}^F   s_{jq}^{(l-1),f} \right)}_{s_{jq}^{(l-1)}}e^{-\frac{t_{i}^{(l),k} - T_q^{(l)}}{\tau_I}}\theta(t_{i}^{(l),k} - T_q^{(l)})\right] K(t-t_{i}^{(l),k}) \nonumber \\
     &~~~~~~~~~~~~~~~~~~~~~~~+  \sum_{j=1}^{N^{(l-1)}} w_{ij}^{(l)}\sum_{q=0}^m \underbrace{\left(\sum_{f=1}^F  s_{jq}^{(l-1),f} \right)}_{s_{jq}^{(l-1)}} \theta(t-T_q^{(l)}) \theta\left(T_q^{(l)} - t_{i}^{(l),k}\right) K\left( t - T_q^{(l)} \right), \\
     &= \left(\sum_{q=0}^m \underbrace{\left(\sum_{j=1}^{N^{(l-1)}} w_{ij}^{(l)}  s_{jq}^{(l-1)}\right)}_{\hat{w}_{iq}^{(l)}} e^{-\frac{t_{i}^{(l),k} - T_q^{(l)}}{\tau_I}}\theta(t_{i}^{(l),k} - T_q^{(l)})\right) K(t-t_{i}^{(l),k}) \nonumber \\
     &~~~~~~~~~~~~~~~~~~~~~~~+  \sum_{q=0}^m \underbrace{\left(\sum_{j=1}^{N^{(l-1)}} w_{ij}^{(l)}  s_{jq}^{(l-1)}\right)}_{\hat{w}_{iq}^{(l)}}\theta(t-T_q^{(l)}) \theta\left(T_q^{(l)} - t_{i}^{(l),k}\right) K\left( t - T_q^{(l)} \right), \\
     &= \left(\sum_{q=0}^{m} \hat{w}_{iq}^{(l)}  e^{-\frac{t_{i}^{(l),k}-T_q^{(l)}}{\tau_I}} \theta(t_{i}^{(l),k} - T_q^{(l)})\right)K\left(t-t_{i}^{(l),k}\right) + \sum_{q=0}^{m} \hat{w}_{iq}^{(l)}\theta(T_q^{(l)} - t_{i}^{(l),k}) K\left(t-T_q^{(l)}\right) 
\end{align}
The subsequent computation can be performed in the same manner as in the TTFS case to obtain the firing time $\hat{t}^{(l),k+1}$.
However, because firing times corresponding to different spike counts cannot be computed in parallel, they must be computed sequentially for each firing event \cite{Yamamoto2024can}.

\section{Experimental details} \label{ss:experiments}

\subsection{Learning algorithm} \label{ss:learning_algorithm}

Training was performed by minimizing the following cost function.
For simplicity, the loss for a single data sample is shown here, although the mini-batch average was used in the actual training:
\begin{align}
\mathcal{C}\left(\{t^{(l)}\}_{l=1,2,\dots, L} \middle| ~\kappa \right)  &= \mathcal{L} \left(t^{(L)} \middle|~\kappa \right) +\mathcal{T}\left(\{t^{(l)}\}_{l=1,2,\dots, L}  \right) \label{eq:cost_function} + \gamma_Q Q.
\end{align}
Here, $t^{(l)}=(t_1^{(l)},t_2^{(l)},\dots,t_{N^{(l)}}^{(l)})$, and $\kappa$ denotes the target vector.
The loss function $\mathcal{L}(\cdot)$ was defined as the following cross-entropy loss \cite{Sakemi2023supervised}:
\begin{align}
\mathcal{L}\left(t^{(L)}\middle|~\kappa\right)  &= -\sum _{i=1}^{N^{(L)}} \kappa _i \ln S_i \left(t^{(L)}\right) \\
S_i\left(t^{(L)}\right)&= \frac{\exp\left(-\frac{t_i^{(L)}}{\tau _\text{soft}}\right)}{\sum_{j=1}^{N^{(L)}} \exp \left(-\frac{t_j^{(L)}}{\tau _\text{soft}}\right)} .
\end{align}
The parameter $\tau_\text{soft}$ is a hyperparameter that scales the softmax function.
The term $\mathcal{T}(\cdot)$ is a temporal penalty term introduced to realize synfire-chain dynamics by restricting the firing times in each layer to a prescribed temporal window.
In Ref. \cite{Sakemi2023supervised}, this penalty was used only to control the output layer.
Here, we extended it so that the firing times in the hidden layers can also be controlled.
In this study, it was defined as
\begin{align}
\mathcal{T}\left(\{t^{(l)}\}_{l=1,2,\dots, L}  \right) &= \sum_{l=1}^L\sum _{i=1}^{N^{(l)}}\left[ \gamma_\text{head} \left(t_i^{(l)} - t^{(l)}_\text{head}\right) ^2   \theta \left(t^{(l)}_\text{head} - t_i^{(l)}  \right) + \gamma_\text{tail} \left(t_i^{(l)} - t^{(l)}_\text{tail}\right) ^2   \theta \left(t_i^{(l)} - t^{(l)}_\text{tail}   \right)\right] \label{eq:temporal_penalty}
\end{align}
Here, $t_\text{head}^{(l)}$ and $t_\text{tail}^{(l)}$ were defined as
\begin{align}
    t_\text{head}^{(l)} = l \cdot \tau_\text{shift} + t_\text{head},~
    t_\text{tail}^{(l)} =  l \cdot \tau_\text{shift} + \tau_\text{width} -  t _\text{tail}.
\end{align}
The parameters $t_\text{head}$, $t_\text{tail}$, $\tau_\text{shift}$, and $\tau_\text{width}$ are hyperparameters.
The term $Q$ was introduced to promote firing of neurons that never fire, for which the temporal penalty $\mathcal{T}$ provides no gradient.  
When non-firing neurons exist, this term drives learning so that the corresponding weights are increased.
It was defined as
\begin{align}
    Q = - \sum_{i,l} \mathbf{1}_{\{t_i^{(l)}=\infty\}} \sum_j w_{ij}^{(l)}. \label{eq:dead_neuron_penalty}
\end{align}
For convolutional layers, the values of those penalties (Eqs. \eqref{eq:temporal_penalty} and \eqref{eq:dead_neuron_penalty}) were divided by the size of the spatial dimensions.

Training was performed by minimizing the above cost function using a gradient-based method.
The weights were updated using AdamW \cite{Loshchilov2017decoupled} with the learning rate $\eta_0$ and the weight regularization coefficient $\gamma_w$.
We observed that training could fail at an early stage when the learning rate was too large.
Therefore, we adopted a warm-up strategy in which the learning rate was set to a small value at the beginning of training and then gradually increased.
This strategy enabled stable training even with a relatively high learning rate.
Let $\eta(p)$ denote the learning rate at epoch $p \in {0, 1, \dots, N^\text{epoch}-1 }$.
The learning rate was updated as follows:
\begin{align}
    \eta (p+1) = \begin{cases}
        2 \eta (p), \text{ for } p < N^\text{warm}, \\
        \eta (p), \text{ otherwise,}
    \end{cases} \eta(0)=\frac{\eta_0}{2^{N^\text{warm}}},
\end{align}
where $N^\text{warm}\in \mathbb{Z}_{\ge0}$ denotes the number of warm-up epochs.
When the warm-up strategy was not used, we set $N^\text{warm}=0$.

In Syn-SNNs, the operation time of each layer is limited to a finite temporal window.
Therefore, the firing dynamics can be accurately approximated with a small number of DSTD steps $M$ by restricting the DSTD time points to the corresponding temporal window.
Specifically, the DSTD parameters were set as $T_\text{first}^{(l)}=l\cdot\tau_\text{shift}-\tau_\text{margin}$ and $T_\text{end}^{(l)}=l\cdot\tau_\text{shift}+ \tau_\text{width} + \tau_\text{margin}$, and $\Delta t=(T_\text{end}^{(l)}-T_\text{first}^{(l)})/M$ was defined accordingly.
In the experiments, we set $\tau_\text{margin}=0.1$.

\subsection{Dataset}

The Fashion-MNIST \cite{Xiao2017fashion} and CIFAR-10 \cite{CIFAR10} datasets consist of two-dimensional image data.
In the Fashion-MNIST dataset, each image has a single channel, whereas the CIFAR-10 dataset consists of three-channel images.
To process these image data, we first normalized the pixel intensities to the range $[0, 1]$.
We then converted the normalized pixel intensities into input spike times as follows:
\begin{flalign}
t_{ijk}^{(0)} = \tau_\text{in} (1-x_{ijk}),
\end{flalign}
where $x_{ijk}$ denotes the normalized pixel intensity, the first and second indices represent the spatial coordinates of the pixel, and the third index represents the channel index ($k\in\{0,1,2\}$, for the CIFAR-10 dataset).
Here, $\tau_\text{in}$ is a positive constant.
In all experiments, we set $\tau_\text{in}=1$.
When spikes were fed into a fully connected layer, the input tensor was reshaped into a one-dimensional tensor.
For the CIFAR-10 dataset, to avoid the problem in which neurons in the first hidden layer fire too early and ignore subsequent inputs, we doubled the number of channels as follows \cite{Sakemi2023sparse}: 
\begin{flalign}
x_{i,j,k} = 1 - x_{i,j,k-3}~(k=3,4,5).
\end{flalign}
Furthermore, following previous work \cite{Zhou2021temporal,Sakemi2023sparse}, we used data augmentation, including horizontal flipping, rotation, and cropping.

\subsection{Network architecture} \label{ss:architectures}

\begin{figure*}
\centering
\includegraphics[clip, width=0.8\textwidth]{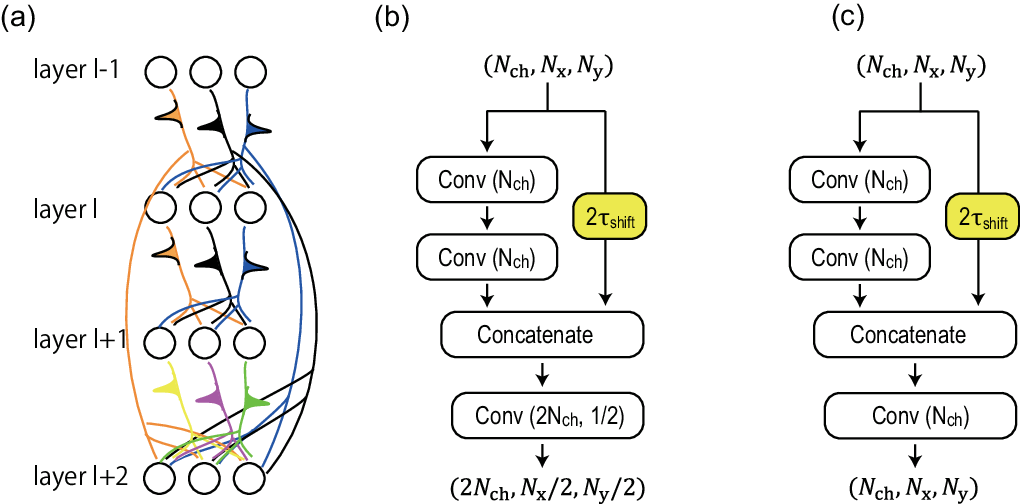}
\caption{{\bf Skip-and-delay block. }
(a) The skip-and-delay (SD) block provides a mechanism for directly projecting output spikes to deeper layers.
Spikes projected from layer $l-1$ are fed into layer $l$ in the usual manner, while the same spikes are also fed into layer $l+2$.
(b) In the SD$1$ block, input spikes to the skip-and-delay block, with channel width $N_\text{ch}$, height $N_\text{x}$, and width $N_\text{y}$, are projected to the neurons in the first and third layers of the block.
The input to the third layer is delayed by $2\tau_\text{shift}$.
The input to the third layer consists of the spikes entering the block and the spikes from the second layer.
The third layer is a convolutional layer with stride 2, which doubles the number of channels and halves the spatial dimensions of the data.
(c) In the SD$_2$ block, all convolutional layers have stride 1.
Therefore, the block outputs spikes with the same number of channels and the same spatial dimensions as the input to the block.
}
\label{fig:skip_connection}
\end{figure*}

In the inference (test) phase of Syn-SNNs, an external reset signal is applied as
\begin{align}
    v_i^{(l)}(l \cdot \tau_\text{shift} + \tau_\text{width}) &\leftarrow 0, \\
    I_i^{(l)}(l \cdot \tau_\text{shift} + \tau_\text{width})&\leftarrow 0.
\end{align}
After the reset signal is applied, the processing of the next data sample, represented by spikes, can start immediately.
In other words, the network realizes pipeline operation, where the next input can be injected before the output of the current input is obtained.
This enables the network to process multiple data samples without degrading the throughput, even when the number of layers is increased.

\begin{table} 
\begin{center}
\caption{Network architecture. C(N,K,S): convolution layer with $N$ output channels, kernel size $K\times K$, and stride $S$. SD$_{1/2}$(N): SD$_{1/2}$ block with $N$ channel size. }\begin{tabular}{lllllll}
Model & Network  \\ \hline
Syn-SNN-9 & C(N,3,1)-C(N,3,1)-C(N,3,1)-C(N,3,2)-C(2N,3,1)-C(2N,3,1)-C(2N,3,2)-C(2N,5,1)-10 \\
Syn-SNN-20 & C(N,K,S)-SD$_2$(N)-SD$_1$(N)-SD$_2$(2N)-SD$_1$(2N)-SD$_2$(4N)-SD$_1$(4N)-10
\end{tabular}
\label{tab:architectures} 
\end{center}
\end{table}

In the experiments, we trained the two network architectures shown in Table \ref{tab:architectures}.
The Syn-SNN architectures are designed following the VGG architecture \cite{Simonyan2015deep} and ResNet \cite{He2016deep}, which are standard architectures for image classification.

Skip connections are known to be effective for efficiently training deep neural networks \cite{He2016deep}.
However, unlike skip connections in ANNs, such as ResNet \cite{He2016deep}, subtraction or addition of output values, namely spikes, is not straightforward for spiking neurons, at least in analog hardware.
Therefore, in this study, we introduce the skip-and-delay block shown in Fig. \ref{fig:skip_connection}.
In the skip-and-delay block, the input from the previous layer is fed not only into the first layer of the block but also into the third layer of the block with a delay of $2\tau_\text{shift}$.
Unlike in previous work \cite{Kim2024rethinking}, the delay is not learned.
Instead, it is fixed to $2\tau_\text{shift}$, corresponding to the delay of two layers in the synfire-chain dynamics.

\subsection{Pareto optimization} \label{ss:Pareto}

\begin{table} 
\begin{center}
\caption{Search space for Pareto optimization. For convenience, fixed parameters are also listed. }
\begin{tabular}{lll}
\hline
parameter & explanation & range \\ \hline
$\eta$    & learning rate & $[10^{-5}, 10^{-2}]$ \\
$\gamma_w$ & weight decay & $[10^{-4},10^{-1}]$ \\
$N^\text{warm}$ & warm-up & $\{0,1,2,3,4\}$ \\
$w_p$  & initial weight scale (positive) & $[1.,20]$ \\
$w_n$ & initial weight scale (negative scale) & $[-0.9, -0.5]$ \\
$\tau_\text{width}$ & time-window width & $[1, 1.5]$ \\
$\tau_\text{shift}$ & time shift of synfire chain & $[0.0, 0.5]$ \\
$\tau $ & time constant of LIF model & $[0.1, 1000]$ \\
$\gamma_\text{head}$ & temporal penalty on head & $[0.01, 0.1]$  \\
$\gamma_\text{tail}$ & temporal penalty on tail & $[0.001, 0.1]$  \\
$\gamma_Q$ & weight-sum penalty & $[10^{-5},10^{-1}]$ \\
$\tau_\text{soft}$ & softmax scale & $[10^{-3},10^{-1}]$ \\
$t_\text{head}$ & temporal penalty's head time & $[0., 0.8]$ \\
$t_\text{tail}$ & temporal penalty's tail time & $[0., 0.5]$ \\
batch size & batch size & 50 \\
$N^\text{epoch}$ & number of epochs & 300 \\
$\tau_\text{margin}$ & DSTD margin & 0.1 \\
$M^\text{test}$ & number of DSTD steps in test phase & 40 \\
\hline
\end{tabular}
\label{tab:search_space} 
\end{center}
\end{table}

In this study, we examined the trade-off between the operating time of each layer, $\tau_\text{width}$, and the classification accuracy in Syn-SNNs.
To perform this analysis, we applied Bayesian optimization to the model parameters and training parameters of Syn-SNNs.
Bayesian optimization was performed using Optuna \cite{Akiba2019optuna}, and approximately 200 trials were conducted for each condition.
The search space used in the optimization is summarized in Table \ref{tab:search_space}.

\section{Benchmarking} \label{ss:benchmarking}

\begin{table} 
\begin{center}
\caption{Continuous-time SNN models. \textit{Method} denotes the method used for gradient computation. ANN: two-phase systems are used to make the model equivalent to ANNs. Adjoint: adjoint equations are used. QIF: a quadratic integrate-and-fire model is used to avoid discontinuities. TTFS: analytical equations of spike timing are used under TTFS coding. Multi-spikes: analytical equations of spike timing are used without the TTFS restriction. F-MNIST: Fashion-MNIST; SHD: Spiking Heidelberg Digits dataset; YY: Yinyang dataset. $^{1}$: nonidealities represented as reversal potentials are included. $^{2}$: accuracies are obtained from figures, $^{3}$: A GPU cluster is used.}
\begin{tabular}{lccccccc}
\hline 
 Model & Method & Architecture & ($\tau_v$, $\tau_I$) &  Pipeline & Accuracy  \\ \hline
 Park et al. \cite{Park2020t2fsnn} & ANN & 17-layer CNN & $(\infty, \tau)$ & YES &  91.43 \% (CIFAR-10) \\
 Sakemi et al. \cite{Sakemi2022spiking} & ANN$^1$ & 784-200-200-10 & ($\infty, \infty$) &  YES & 98.46 \% (MNIST) \\
Stanojevic et al. \cite{Stanojevic2024high} & ANN & 17-layer CNN & ($\infty, \infty$) & YES & 93.69 \% (CIFAR-10)  \\
 Sakemi et al. \cite{Sakemi2025harnessing} & ANN$^1$ & 8-layer CNN & ($\infty, \infty$) & YES & 88.5 \% (CIFAR-10) \\ \hline
 Wunderlich et al. \cite{Wunderlich2021event} & adjoint & 784-200-10 & ($4\tau$, $\tau$) & NO & 97.6 \% (MNIST) \\
 Meszaros et al. \cite{Meszaros2025efficient} & adjoint & 700-1024-1024-20 & ($4\tau, \tau$)  & NO & 78.5 \% (SHD)$^2$ \\
 Konig et al. \cite{Konig2026eventax} & adjoint & 784-200-10 & ($4\tau, \tau$) & NO & 97.5 \% (MNIST) \\ 
 Klos et al. \cite{Klos2025smooth} & QIF & 784-100-100-10 & N/A & NO & 97.3 \% (MNIST) \\
 Wenig et al. \cite{Wenig2026quadratic} & QIF & 700-128-20 & N/A & NO & 90.1 \% (SHD)\\
 Goltz et al. \cite{Goltz2025DelGrad} & TTFS & 4-30-3 & ($2\tau, 2\tau$) & NO & 92.6 \% (YY) \\
  Mostafa \cite{Mostafa2018supervised} & TTFS & 784-800-10 & ($\infty, \tau$) & NO & 97.55 \% (MNIST)\\
  Comsa et al. \cite{Comsa2021temporal} & TTFS & 786-340-10 & ($\tau, \tau)$ & NO & 97.96 \% (MNIST) \\
  Goltz et al. \cite{Goltz2021fast} & TTFS & 256-246-10 & ($2\tau$, $\tau$) & NO & 97.4 \% (MNIST)  \\
 Sakemi et al. \cite{Sakemi2023supervised} & TTFS & 784-800-10 & ($\infty, \infty$) & NO & 98.34 \% (MNIST)\\
 Zhang et al. \cite{Zhang2021rectified} & TTFS &  6-layer CNN & ($\infty, \infty)$ & NO & 99.4 \% (MNIST) \\
 Zhou et al. \cite{Zhou2021temporal} & TTFS & 16-layer CNN & ($\infty$, $\tau$) &  NO & 92.68\% (CIFAR-10)$^3$\\
Sakemi et al. \cite{Sakemi2023sparse} & TTFS & 5-layer CNN & ($\infty$, $\infty$) &  NO & 79.26\% (CIFAR-10) \\
 Yamamoto et al. \cite{Yamamoto2024can} & multi-spikes & 784-400-10 & ($2\tau, \tau)$ & NO & 98.43 \% (MNIST) \\
 Bacho et al. \cite{Bacho2023exploring} & multi-spikes & 784-800-10 & ($2\tau, \tau)$ & NO & 98.88 \% (MNIST) \\
 Morrill et al. \cite{Morrill2026bullet} & multi-spikes & 700-512-512-20 & ($4\tau, \tau$) & NO & 94.96 \% (SHD)  \\
Syn-SNN-20 (proposed) & TTFS (DSTD) & 20-layer CNN & ($\infty, \infty$) & YES &  92.33 \% (F-MNIST) \\
Syn-SNN-9 (proposed) & TTFS (DSTD) & 9-layer CNN & ($\infty$, $\infty$) & YES & 90.36 \% (CIFAR-10)\\ 
\hline

\end{tabular}
\label{tab:continuous_SNN} 
\end{center}
\end{table}

Training multilayer continuous-time SNNs is generally challenging, and consequently, the number of reported studies remains much more limited than that for discrete-time SNNs \cite{Dampfhoffer2023backpropagation}.
Table \ref{tab:continuous_SNN} summarizes recently reported continuous-time SNN models.

\section{Additional experiments} \label{ss:additional_experiments}

\subsection{Convergence of DSTD approximations} \label{ss:DSTD_convergence}

\begin{figure*}
\centering
\includegraphics[clip, width=1.0\textwidth]{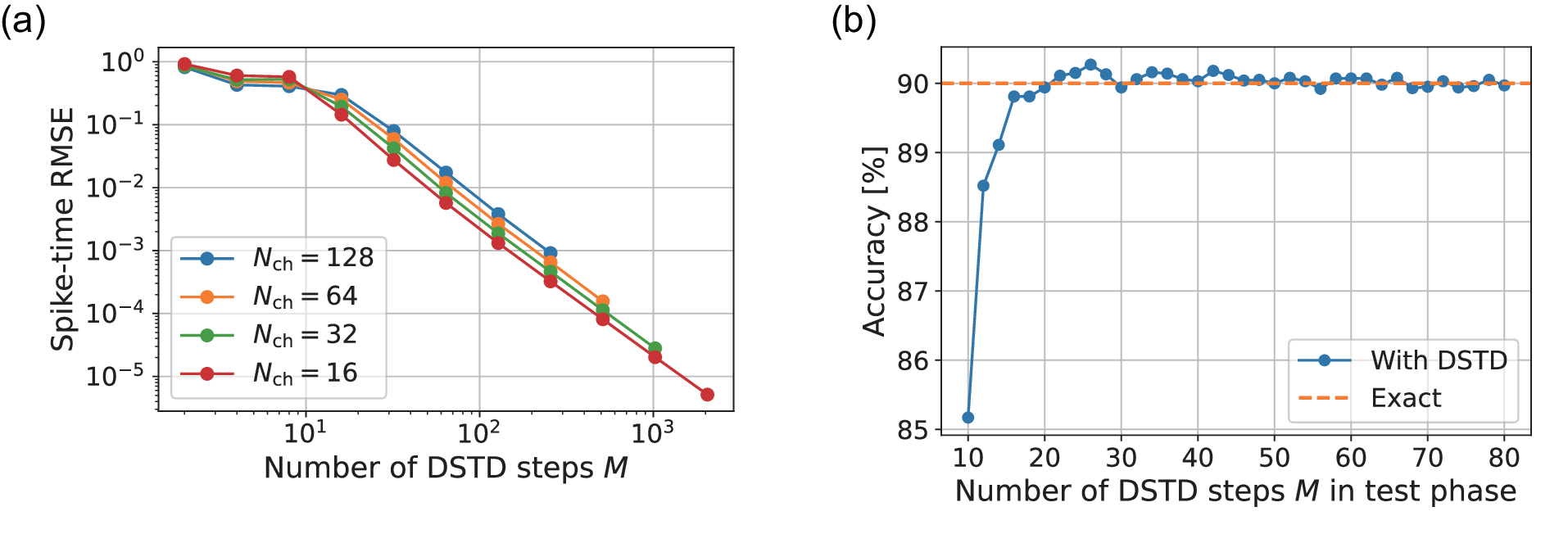}
\caption{{\bf Convergence analysis. }
(a) Root mean squared errors between the exact spike times $t_i^{(L)}$ in the output layer and the approximated spike times $\hat{t}_i^{(L)}$ when the number of DSTD steps $M$ is increased in untrained nine-layer convolutional SNNs. Results are shown for networks with different widths $N_\text{ch}$.
(b) Classification accuracy on the CIFAR-10 test data for different numbers of DSTD steps $M$ in a trained nine-layer convolutional SNN. The orange dashed line indicates the classification accuracy obtained using the exact-solution model.
}
\label{fig:convergence}
\end{figure*}

The approximation error of DSTD can be reduced by increasing the number of steps \cite{Sakemi2025harnessing}.
Here we empirically demonstrate its convergence.
Figure \ref{fig:convergence} (a) shows the approximation error of the output-layer spike times in nine-layer convolutional SNNs.
The approximation error is defined as follows:
\begin{align}
E_\text{RMSE} = \sqrt{\frac{\sum_{d=1}^{D} \sum_{i=1}^{N^{(L)}} \left( t_i^{(L),d} - \hat{t}_i^{(L),d} \right)^2}{DN^{(L)}}},
\end{align}
where $D$ is the number of data samples, $N^{(L)}$ is the number of neurons in the final layer, $t_i^{(L),d}$ is the firing time of a final-layer neuron obtained from the exact solution in the $d$th data sample, and $\hat{t}_i^{(L),d}$ is the firing time approximated by DSTD in the $d$th data sample.
We randomly sampled 100 examples from the CIFAR-10 test data ($D=100$) and evaluated the approximation error for the output layer with 10 neurons ($N^{(L)}=10$).
When DSTD was used, the DSTD-based approximation was applied to all layers.
The networks were evaluated at their initial-weight states, and the experiments were performed with $\tau_\text{width}=2$.
The results show that the approximation error decreases monotonically when $M$ becomes sufficiently large.

Furthermore, Fig. \ref{fig:convergence} (b) shows the classification accuracy on the CIFAR-10 test dataset for a trained nine-layer convolutional SNN using different numbers of DSTD steps $M$.
The accuracy obtained using the exact solution is also shown by the orange dashed line.
These results indicate that $M=40$ DSTD steps are sufficient to achieve accuracy close to that of the exact-solution model.

\bibliographystyle{unsrt}
\bibliography{myBib}

\end{document}